%% file: emnlp2023.tex
\pdfoutput=1

\documentclass[11pt]{article}

\usepackage[]{EMNLP2023}
\usepackage{times}
\usepackage{latexsym}

\usepackage[T1]{fontenc}

\usepackage[utf8]{inputenc}

\usepackage{microtype}

\usepackage{inconsolata}

\usepackage{amsmath}
\usepackage{amssymb}
\usepackage{mathtools}
\usepackage{amsthm}
\usepackage{booktabs} 
\usepackage{multirow} 
\usepackage{geometry} 
\usepackage[ruled,vlined, linesnumbered]{algorithm2e}
\usepackage{float} 
\usepackage{paralist} 
\usepackage{xspace}
\title{Beyond Label Attention: Transparency in Language Models for Automated Medical Coding via Dictionary Learning}

\author{John Wu \\ University of Illinois Urbana-Champaign \\ \texttt{johnwu3@illinois.edu}
        \AND
        David Wu \\ Vanderbilt University \\ \texttt{David.h.wu@vanderbilt.edu}
        \And
        Jimeng Sun \\ University of Illinois Urbana-Champaign \\ \texttt{jimeng@illinois.edu}}

\begin{document}
\maketitle
\begin{abstract}
Medical coding, the translation of unstructured clinical text into standardized medical codes, is a crucial but time-consuming healthcare practice. Though large language models (LLM) could automate the coding process and improve the efficiency of such tasks, {\em interpretability} remains paramount for maintaining patient trust. Current efforts in interpretability of medical coding applications rely heavily on label attention mechanisms, which often leads to the highlighting of extraneous tokens irrelevant to the ICD code. To facilitate accurate interpretability in medical language models, this paper leverages {\em dictionary learning} that can efficiently extract sparsely activated representations from dense language model embeddings in superposition. Compared with common label attention mechanisms, our model goes beyond token-level representations by building an interpretable dictionary which enhances the mechanistic-based explanations for each ICD code prediction, even when the highlighted tokens are medically irrelevant. We show that dictionary features can steer model behavior, elucidate the hidden meanings of upwards of 90\% of medically irrelevant tokens, and are human interpretable.
\end{abstract}

\footnote{Code available at: \url{https://github.com/jhnwu3/BeyondLabelAttention}}
\input{sections/intro}

\input{sections/related_work}
\input{sections/method}

\input{sections/results}

\input{sections/conclusion}

\section*{Limitations}
We note that there are several key limitations of the dictionary learning applied here. First, as noted in the training details within the appendix, the sparse autoencoders are unable to perfectly reconstruct PLM embeddings, and thus cannot fully capture a model's downstream performance \cite{cunningham2023sparse, bricken2023monosemanticity, templeton2024scaling_mono}. Furthermore, we note as similarly explored by \cite{cunningham2023sparse, bricken2023monosemanticity, templeton2024scaling_mono} that there exist dead or missing features, features that do not activate regardless of token embedding or concepts that exist in the model but are not captured within the dictionary, indicating not all features are useful or interpretable.

In terms of the coherence of highly activating features, we highlight that the sparse dictionaries learned are not as coherent as supervised mappings of the LAAT, limiting its human understandability. Furthermore, we observe that simply interpreting the PLM embeddings at the last layer, which \cite{yun2023transformer_bert_visualization_dictionary, cunningham2023sparse} claim to be the most interpretable, is insufficient for fully capturing and interpreting a PLM's behavior. Referring to Table \ref{table: Model Steering}, the resolution of our dictionary features is lacking in comparison to the granularity of medical codes, where we observe that the total number of highly meaningful features (i.e., those that can change a medical code's prediction) is smaller than the number of medical codes highly affected, suggesting that the features learned most likely represent higher abstract concepts than highly-specific medical codes.

Recently, \cite{templeton2024scaling_mono} has explored scaling up sparse autoencoders to potentially millions of dictionary features for extremely large language models. They show that not all semantic features are learnable, especially in cases without access to large amounts of diverse data \cite{bricken2023monosemanticity}. While synthetic data generation exists to further augment the diversity of tokens in the dictionary learning corpus, the size of sparse autoencoders must also increase as model complexity increases \cite{templeton2024scaling_mono} in order to maintain the granularity of dictionary features. Regardless, such mechanistic explanations are still fairly efficient when compared to black-box alternatives. To the best of our knowledge, its applicability to smaller scale language datasets is unknown.

While our experiments revealed several key benefits of using sparse autoencoders for interpretability, future work offers exciting avenues to push medical PLM interpretability further. Despite SPINE's \cite{subramanian2017spine} improvements over vanilla $L_1$ methods, sparse autoencoders' reconstruction limitations can restrict their ability to capture all information relevant to downstream predictions. Exploring more expressive representation learning techniques like causal and disentanglement methods \cite{schölkopf2021causal, wang2023disentangled} could hold promise for interpretability gains. Additionally, unraveling the complex relationships within medical PLMs through both automated circuit discovery \cite{conmy2023automated_circuit} and dictionary learning approaches \cite{cunningham2023sparse} could offer further valuable insights into an ICD code prediction.

\section*{Ethics Statement}
We note that all clinical notes used from the MIMIC-III dataset \cite{Johnson2023MIMIC4, Johnson2016} are deidentified and that our method is heavily focused on medical tokens and their conceptual meanings towards ICD predictions. Thus, we follow guidelines laid out by PhysioNet's MIMIC3 health data license \cite{Johnson2016} and note our study does not contain any additional patient information that can lead to privacy violations. We note that our expert human annotators have consented and are our collaborators. 


\bibliography{anthology,custom}
\bibliographystyle{acl_natbib}

\appendix
\input{sections/appendix}

\end{document}

%% file: sections/intro.tex
\section{Introduction} \label{sec:intro}
Transparency is a vital factor in healthcare to gain patients' trust, especially when AI models make critical decisions in clinical practice \cite{Rao2022-kj_TransparencyMedicalCare}. One of the essential applications of AI models is to assign International Classification of Diseases (ICD) codes automatically based on the clinical text (we name this task as {\em medical coding}). These ICD codes categorize patient diagnoses, conditions, and treatments for billing, reporting, and treatment purposes \cite{HistoryOfICD, Johnson2021-aw-ICD-coding-injury}. However, assigning ICD codes is complex and requires expertise and time \cite{OMalley2005-we-icd-coding-hard}. Recent advancements in medical pre-trained language models (PLMs) have made it possible to treat medical coding as a high-dimensional multilabel classification challenge \cite{AutomatedMedicalCoding, huang2022plmicd}. These AI models led to significant success in efficient ICD coding \cite{kaur2021systematic_review}.  
However, their transparency remains of great concern \cite{HAKKOUM2022108391_medical_interpretability_review}.
Therefore, developing automated interpretability methods is crucial to upholding transparency in medical coding processes.

Significant progress has occurred in the field of black-box interpretability, particularly concerning feature attribution, with the emergence of perturbation-based methods such as SHAP \cite{lundberg2017unifiedshap} and its approximate counterpart LIME \cite{ribeiro2016whyLime, moraffah2020causal_lime}. These techniques, rooted in information and game theory, are recognized for assessing feature relevance in detail by intelligently perturbing and ablating input features \cite{lundberg2017unifiedshap, ribeiro2016whyLime}. While approximation methods have greatly improved the speed of calculating Shapley values, exact computations remain expensive \cite{Lundberg2020ShapLocalStuffForTrees, Chen2022SeriesOfModels, shrikumar2019learningDeepLift, mosca-etal-2022-shap-NLP}. The huge computational cost makes their application impractical towards automated medical ICD coding since clinical notes usually contain thousands of high dimensional token embeddings in a vast multilabel prediction space \cite{Johnson2023MIMIC4}. 

{\em As such, we seek a human-interpretable approach that scales efficiently with large datasets, highlights essential features, and offers more comprehensive explanations of PLM predictions.} Recent advancements in mechanistic interpretability methods \cite{cunningham2023sparse, räuker2023transparent_survey_mechanistic} have demonstrated the potential to surpass the computational challenges posed by traditional black-box approaches by elucidating the roles of specific neuron subsets within a network. This level of mechanistic understanding is precious in the medical field, where explaining the significance of a feature is as crucial as its identification. 

In recent years, the attention mechanism \cite{abnar2020quantifyingattention, vaswani2023attentionIsAllYouNeed, chefer2021generic, chefer2022optimizing} has been heavily used to interpret and explain the behavior of large transformer models. Within the realm of automated medical ICD coding, label attention (LAAT) variants are most prevalent due to their efficiency in handling extensive sequence lengths. They calculate an attention score for each token relative to each label, thereby identifying tokens crucial for ICD code predictions \cite{mullenbach2018explainable_CAML,Vu_2020_LAAT}. Nevertheless, studies such as \cite{pandey2023interpretabilityofattentionetworks} have questioned the interpretability and validity of explanations provided by attention mechanisms.

\begin{figure*}[ht]
\centering
\includegraphics[width=1.0\textwidth]{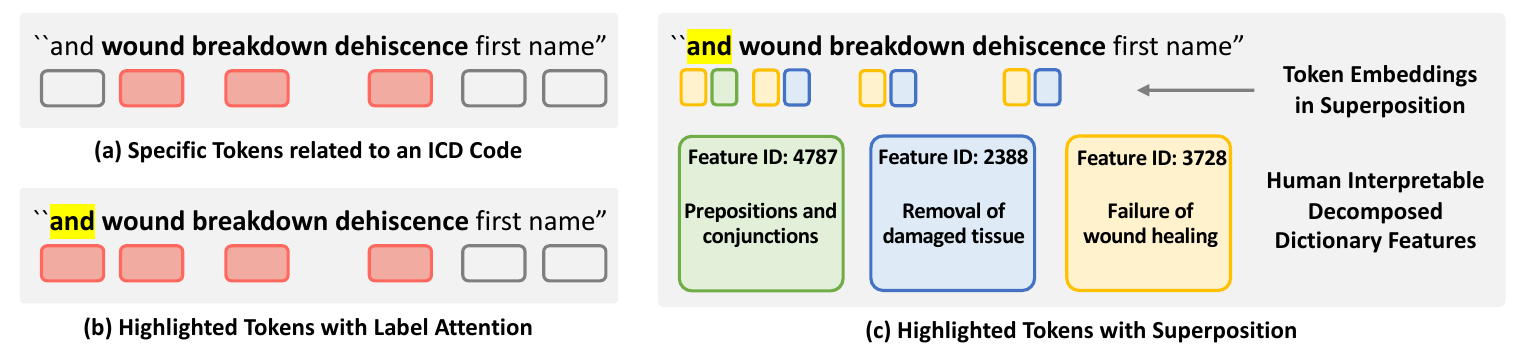}
\caption{Motivation: LAAT identifies the most relevant tokens for each ICD code (b). Compared to our inspection of which tokens are most relevant to an ICD code (a), we assume "and" is irrelevant to an ICD code prediction. Although it may appear as though "and" is irrelevantly highlighted, taking token embeddings out of superposition allows us to decompose dense token embeddings into more semantically meaningful dictionary features that show that concepts of "failure of wound healing" are embedded within its token embedding (c), thereby giving justification for its highlighting by LAAT for a wound-related ICD code. }
\label{fig:motivation}
\end{figure*}

For example, the LAAT mechanism may highlight incoherent or irrelevant tokens, such as stop words for highly medically specific ICD codes. For instance, LAAT attributes the stop token "and" to the medically specific ICD code "998.59 postoperative wound infection" despite conjunctions being irrelevant to medical prognosis, thus undermining the interpretability of LAAT as shown in Figure \ref{fig:motivation}. We attribute this issue to be the result of neuron polysemanticity—where a single neuron responds to diverse, unrelated inputs—complicates direct interpretation \cite{olah2020zoomPolySemanticity}. 

\citet{elhage2022superposition} theorizes polysemanticity to be a form of superposition, occurring when the count of independent data features surpasses layer dimensions, leading to data features being represented by linear combinations of neurons. As a result, individual neurons are often directly uninterpretable \cite{subramanian2017spine, cunningham2023sparse}.  
Addressing this, sparse autoencoders \cite{OLSHAUSEN19973311_OGSparseCoding} have been applied to distill these complex dense representations into interpretable, sparse linear combinations, performing what is known as {\em dictionary learning} (DL). This strategy has effectively decomposed different language model layers, such as neural word embeddings \cite{subramanian2017spine}, MLPs \cite{cunningham2023sparse, bricken2023monosemanticity}, and residual connections \cite{yun2023transformer-residual}, proving scalable and monosemantic. 

Our paper generalizes these sparse coding concepts to better understand the attention mechanism and pre-trained language model (PLM) embeddings by constructing effective dictionaries with LAAT to improve the interpretability of the medical ICD coding task. We summarize our main contributions:

\begin{itemize}
    \setlength{\itemsep}{0pt}
    \setlength{\parskip}{0pt}
    \setlength{\parsep}{0pt}
    \item Expanding on \cite{bricken2023monosemanticity}'s ablation studies, we show that combining learned dictionary features with another mechanistic component (LAAT) \cite{Vu_2020_LAAT}, in our new interpretability framework AutoCodeDL, improves explainability of downstream ICD predictions.
    \item We build medically relevant dictionaries with sparse autoencoders that can capture medically relevant concepts hidden within superposition.
    \item We develop new automated proxy metrics for assessing the human understandability of constructed dictionaries and conduct extensive evaluations on large scale clinical text-based corpus.
\end{itemize}

%% file: sections/related_work.tex
\vspace{-0.2cm} 
\section{Related Work} 
\vspace{-0.1cm} 
\subsection{Automated Interpretability in ICD Coding}
Alternative automated ICD coding methods, like phrase matching \cite{cao-etal-2020-clinical_ICD_phrase_matching} and relevant phrase extraction using manually curated knowledge bases \cite{DUQUE2021102177_Knowledge_base}, offer inherent interpretability but fall short in expressive power compared to neural network-based approaches. This discrepancy highlights a persistent tradeoff between interpretability and performance in ICD coding tasks. 

Furthermore, the prevailing interpretability method for deep neural models in ICD coding tasks rely on the attention mechanism \cite{YAN2022161SurveyICDCoding}. Specifically, the LAAT mechanism projects token embeddings into a label-specific attention space, where each token receives a score indicating its relevance to each ICD prediction. Such attention-based associations between tokens and classes have been employed in various architectures, including convolutional models like CAML \cite{mullenbach2018explainable_CAML}, recurrent neural networks \cite{Vu_2020_LAAT}, and large language models \cite{huang2022plmicd, yang2023surpassingGPT4}. While computationally efficient, it overlooks the potentially richer information hidden within the embedding space, hindering our understanding of automated ICD predictions. Our work builds on top of such past works, and directly interprets the medical PLM embeddings.

\subsection{Dictionary Learning}
{\em Dictionary learning} aims to find a sparse representation of input data in the form of linear combination of basic elements \cite{OLSHAUSEN19973311_OGSparseCoding}. Such an approach has been applied across various domains, including word embedding decomposition \cite{subramanian2017spine}, interpretation of language model activations \cite{bricken2023monosemanticity, cunningham2023sparse, yun2023transformer-residual}, enhancement of representation learning \cite{ghosh2023dictionary_representations, tang2023dictioanry_trajectories}, and analysis of time-series data \cite{xu2023_dictionary_time_series}, highlighting their versatility \cite{Zhang_2015_survey_sparse}. 

Our research aims to understand if and how dictionary learning improves upon existing interpretability methods like LAAT for predicting medical codes (ICDs) in a highly practical setting. By analyzing diverse medically relevant ICDs, we assess the learned dictionaries' ability to capture specific and meaningful medical concepts. Unlike previous work requiring extensive human annotation \cite{bricken2023monosemanticity, cunningham2023sparse, subramanian2017spine}, we propose new automated metrics to measure how understandable these dictionaries are due to the cost of expert annotation.

%% file: sections/method.tex
\vspace{-0.1cm} 
\section{Methodology} \label{sec:method}
\begin{figure*}[ht] 
\centering
\includegraphics[width=1.0\textwidth]{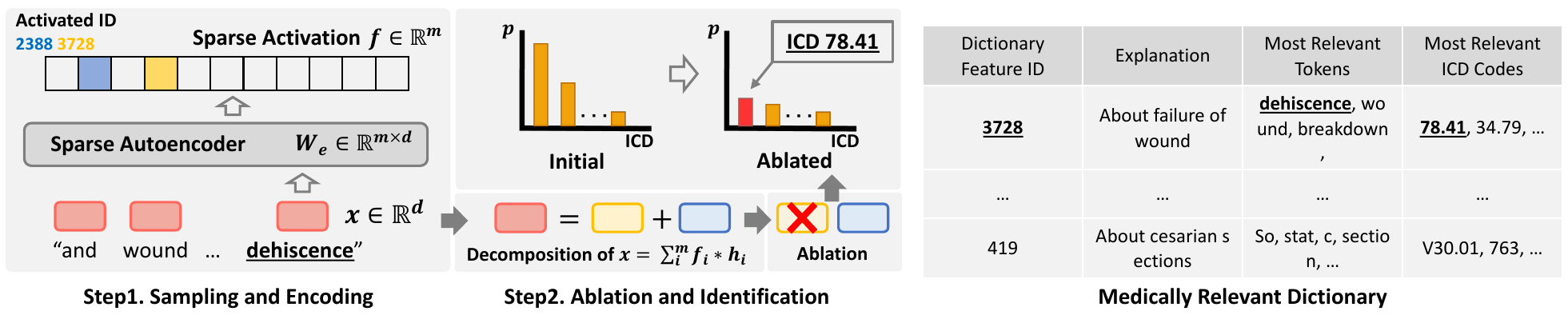}
\caption{Building a dictionary involves several steps: A sparse autoencoder decomposes each token embedding into a sparse latent space, where each nonzero element represents a dictionary feature ID (step 1). This process enables the creation of mappings between tokens and various dictionary features and ICD codes. In step 2, ICD codes are mapped to dictionary features based on the softmax probabilities of each ICD prediction after dictionary embedding ablations, as detailed in section \ref{sec:Method Exp}. Once a dictionary is constructed, it is utilized to enhance explanations by applying it to highlighted tokens identified by LAAT in Figure \ref{fig:proposed_method}.}
\label{fig:method}
\end{figure*}

\vspace{-0.1cm} 
As depicted in Figure \ref{fig:method}, our focus within dictionary learning involves explicitly building dictionaries where relevant tokens and ICD codes are mapped to dictionary features. We examine two sparse autoencoder approaches aimed at creating interpretable representations from dense language model embeddings: one via $L_1$ minimization \cite{cunningham2023sparse, bricken2023monosemanticity} and another via SPINE's loss function \cite{subramanian2017spine}. While our discussion primarily centers on the $L_1$ minimization technique for its widespread application and illustrative clarity regarding dictionary learning's objectives in section \ref{sec:saenc}, further details on SPINE are provided in the Appendix~\ref{appendix:spine}.

Then, using our trained sparse autoencoder, we perform ablation studies to understand the downstream effects of dictionary features in section \ref{sec:ExpResults} and map the relevant ICD codes to each dictionary feature as discussed in section \ref{sec:Method Exp}. Finally, we leverage sparse encoding and its ablation techniques in constructing our final dictionary, mapping both relevant tokens and ICD codes to each dictionary feature in section \ref{sec:BuildingDictionary}, which is used in our new proposed method AutoCodeDL in Figure \ref{fig:proposed_method}.

\subsection{Sparse Autoencoders} \label{sec:saenc}
Following \cite{bricken2023monosemanticity}'s approach, let $x \in \mathbb{R}^d$ be the token embedding we wish to interpret, $d$ the dimension size of the token embedding, and $m$ the dimension size of the latent sparse dictionary feature activations $f \in \mathbb{R}^m$ generated by the sparse autoencoder. Our $L_1$ sparse autoencoder is shown below in equations \ref{eq:L1 Autoencoder}
through \ref{eq: L1 Loss} where $W_e \in \mathbb{R}^{m \times d}$ is the encoder weight matrix, $b_e \in \mathbb{R}^m$ is the encoder weight bias term, $b_d \in \mathbb{R}^d$ is the decoder bias term, and $W_d \in \mathbb{R}^{d \times m}$ represents the sparse dictionary embeddings:
\begin{align}
    \bar{x} &= x - b_d\label{eq:L1 Autoencoder} \\
    f &= \text{ReLU}(W_e \bar{x} + b_e) \\
    \hat{x} &= W_d \cdot f + b_d \\
    \mathcal{L} &= \frac{1}{|X|} \sum_{x \in X} \left\lVert x - \hat{x} \right\rVert_2^2 + \lambda_{L_1} \left\lVert f \right\rVert_1\label{eq: L1 Loss} 
\end{align}

The $L_1$ norm $\left\lVert f \right\rVert_1$ in the loss function (see eq. \ref{eq: L1 Loss}) enforces the sparse representation of $f$ in training. As a result, only certain elements within $f$ activate
for certain types of token embeddings $x$, creating a direct mapping between different tokens (words) represented by $x$ and their respective features identified by $f_i$. For instance, we observe that "depression"-related tokens only activate the dictionary feature $f_{5732}$.

While there exist many other useful properties of $L_1$ minimizations \cite{Zhang_2015_Survey_Sparse_Representations}, the key idea is that a sparse linear combination of learned dictionary embeddings represents every token embedding. First, let us examine the dictionary embedding matrix $W_d$ as defined below. 
\begin{equation}
    W_d = \begin{bmatrix} h_{0}, & h_{1}, & \ldots &, h_{m} \end{bmatrix}^T
\end{equation}

Let $h_{i} \in \mathbb{R}^d $ be the dictionary embedding associated with a dictionary feature $f_i$; we have the following decomposition of any token embedding $x$ into sparse features. 

\begin{equation}
    x \approx \sum_i^m f_i * h_{i}
\end{equation}

Another key intuition behind why these decompositions are directly interpretable is that since certain $f_i$ only activate (i.e., is nonzero) for certain types of tokens, its dictionary embeddings $h_i$ have a defined direction in the PLM embedding space that should directly correspond to some meaningful concept within the original clinical text, thus having downstream implications.

\subsection{Mapping Dictionary Features to ICD Codes} \label{sec:Method Exp}
To build a medically relevant dictionary, ICD codes should map to their respective meaningful dictionary features. Following the methodology in \cite{bricken2023monosemanticity}, we ablate features in clinical notes by targeting any activated dictionary feature $f_i > 0$ in token embedding $x \in \mathbb{R}^d$. For each feature $f_i$ with corresponding feature embedding $h_{i} \in \mathbb{R}^d$, we define the ablated token embedding as $\Tilde{x}$.
\begin{equation} \label{eq:feature ablation}
    \Tilde{x} = x - f_i \cdot h_{i}
\end{equation}

For any token in a clinical note, we perform token embedding ablations, recalculate the ablated model's softmax probabilities for all $\mathbb{C}$ classes or ICD codes, and compute the probability differences $\delta_i$. 
\begin{equation} \label{eq:}
    \delta_i = p(x) - p(\Tilde{x}), \delta_i \in \mathbb{R}^{\mathbb{C}} 
\end{equation}

Finally, for any given class $c \in \{1,2,\ldots,C\}$, we sort and record the top $\delta_{i,c}$'s when ablating each dictionary feature $f_i$ and its embedding $h_i$, identifying its most relevant ICD codes. Such ablations are later used for evaluating the model explainability of dictionary features and building more human-interpretable medical dictionaries with the sparse autoencoder.

\subsection{Building Medically Relevant Dictionaries to Augment ICD Explanations} \label{sec:BuildingDictionary}
In essence, the sparse autoencoder contains a dictionary within its latent space. While efficient in time and space complexity, its direct interpretation requires the construction of a more human-interpretable and literal dictionary containing the most relevant tokens and ICD codes for each dictionary feature $f_i$. Building a dictionary can be summed up into two sorting steps.

\textbf{Sampling and Encoding.} We sample a certain number of clinical notes from our test set. For every token in each clinical note, we encode their PLM embeddings with our sparse autoencoder and retrieve its sparse feature activations $f_i$. Then, for each dictionary feature $i$, we sort by each token's respective $f_i$ and select the top $k$ tokens with the highest feature activations for each dictionary feature. Since there are often compound words, consisting of multiple tokens, we also retrieve any neighboring tokens with nonzero activations. 

\textbf{Ablation and Identification.} For every clinical note, we perform ablations for each activated dictionary feature's embedding $h_i$, measuring the change in predicted probability for each ICD code. For each dictionary feature, we identify its most relevant classes based on the largest probability drops after ablation. We formalize this process for a single clinical note in the pseudocode shown in algorithm \ref{alg:build_dict} in the Appendix.

\begin{figure*}[ht] 
\centering
\includegraphics[width=1.0\textwidth]{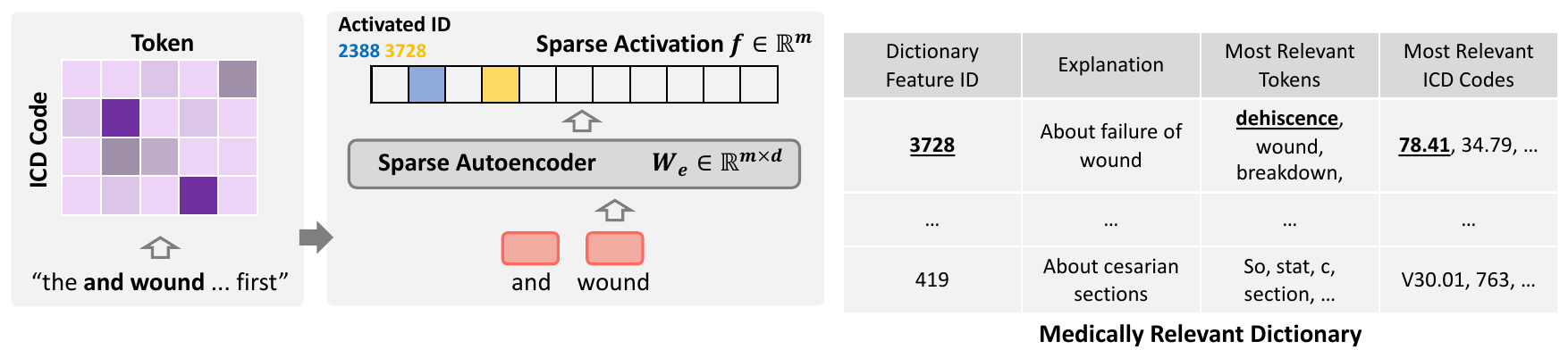}
\caption{Proposed method for automated ICD interpretability pipeline: \textbf{AutoCodeDL}. LAAT identifies the most important words "and wound". Then, the sparse autoencoder queries its most activated dictionary features, returning its respective dictionary feature ids that can be leveraged to further explain the PLM's predictions and attention highlights.}
\label{fig:proposed_method}
\end{figure*}

\textbf{Proposed Method of Interpretability.}
After constructing the dictionary, we can utilize it to query the dictionary features of any PLM token embeddings, enhancing interpretability. Integrated with LAAT in our proposed new method AutoCodeDL, we initially identify the key tokens for each ICD prediction and subsequently match their embeddings to features in our dictionaries, thereby refining explanations (see Figure \ref{fig:proposed_method}).

\vspace{-0.3cm}

%% file: sections/results.tex
\section{Interpretability Evaluations} \label{sec:results}
\vspace{-0.1cm} 
While there does not exist a unified singular definition of neural model interpretability, there is a general consensus that model interpretability methods should be both model explainable and human understandable \cite{Zhang_2021SurveyNNInterpretability}. We define explainability as how much do the dictionary features learned predict the pre-trained language model's ICD  predictions. We define human understandability within the lens of monosemanticity: a dictionary feature is only human interpretable when the tokens that highly activate said dictionary feature are related and its underlying concept can be easily identified.

\textbf{Sparse Autoencoders.} We explore two major dictionary learning (DL) sparse autoencoder approaches, SPINE by \cite{subramanian2017spine} and $L_1$ described in section \ref{sec:method}.

\textbf{Baselines.} Following \cite{cunningham2023sparse}, we explore four unsupervised baseline encoders that we compare to a true dictionary encoding: Independent Component Analysis (ICA), capable of decomposing word embeddings into semantically meaningful independent components (ICs) \cite{musil2022independent_ICA}; Principal Component Analysis (PCA), which has been used to analyze the structure of word embeddings \cite{musil2019examining_words_PCA}; an Identity ReLU encoder, which effectively treats each element in the PLM embedding as its own dictionary feature; and a random encoder.

\textbf{Dataset and Model.} We train on PLM embeddings from a 110M medical RoBERTa model leveraged by the state of the art PLM-ICD coding model \cite{huang2022plmicd, lewis-etal-2020-pretrained-medical-roberta} and evaluate our method using the cleaned MIMIC-III dataset, using 38,427 clinical notes for training and 8,750 for evaluation, as detailed by \cite{AutomatedMedicalCoding}. Additional details on training are in Appendix~\ref{appendix:saenc_training}.

\subsection{Model Explainability} \label{sec:ExpResults}
\textbf{Faithfulness.} 
Inspired by explainability evaluation metrics within the vision domain \cite{GradCamPP, samek2015evaluating_visualization}, we assess our interpretable dictionary features by removing them (ablation) and measuring their impact on predicted ICD codes. This approach helps us quantify how well our explanations align with the model's predictions. For example, ablating the "depression" feature should primarily impact related ICD codes like "depressive disorders" while leaving unrelated ones like "postoperative wound infection" relatively unaffected. 

To address this, we consider both the \emph{decrease} in the most likely code's softmax probability after ablation and the sum of absolute changes in softmax probabilities for all other codes. Computing a ratio of these measures (detailed in Table \ref{tab:ablation}) allows us to accurately gauge an interpretability method's explanatory power. This ablation metric is analogous to Comprehensiveness (Comp) by \cite{chan-etal-2022-comparative_faithfulness, deyoung2020eraser_comprehensiveness}, but instead of erasing entire tokens, we ablate specific components of the embedding (see eq. \ref{eq:feature ablation}). 

\textit{\textbf{Setup.}} We compare our interpretability framework, AutoCodeDL, against three sets of baselines. The first set involves feature ablations using baseline encoders and full token ablations (labeled "token" in Table \ref{tab:ablation}) with LAAT to pre-highlight relevant tokens. The second set excludes LAAT and simply explains only with DL feature ablations across all tokens. The third set excludes LAAT and only utilizes the baseline encoder ablations. For additional details on the ablations for each baseline encoder, please refer to Appendix ~\ref{appendix:baseline ablation details}.
    
\textit{\textbf{Results.}} From Table \ref{tab:ablation}, ablations of dictionary features of highlighted tokens with our proposed AutoCodeDL method show minimal impact on other ICD code predictions while retaining large drops in softmax probabilities. As a result, our proposed method outperforms all baselines in our ratio explainability metric. Within our baselines, their ablations varied tremendously. Most notably, ablating ICs from embeddings using ICA had minimal impact on predictions, possibly because ICA approximations identify features with negligible mutual information while principal component ablations were similar to full token ablations. 

\begin{table*}[h!]
\centering
\resizebox{1.0\textwidth}{!}{%
\begin{tabular}{c|c c |c c c c c | c c |c c c c}
\toprule
\multicolumn{13}{c}{\textbf{Ablating Dictionary Features of Highlighted Tokens}} \\
\midrule
\multicolumn{1}{c|}{\textbf{Experiment}} & \multicolumn{2}{c|}{\textbf{ AutoCodeDL }} & \multicolumn{5}{c|}{\textbf{LAAT + Baselines }} & \multicolumn{2}{c}{\textbf{ DL }}  & \multicolumn{4}{|c}{\textbf{ Baselines }} \\
\midrule
 & \textbf{L1} & \textbf{SPINE} & \textbf{ICA} &  \textbf{PCA} & \textbf{Identity} & \textbf{Random}  & \textbf{Token} & \textbf{L1} & \textbf{SPINE} & \textbf{ICA} &  \textbf{PCA} & \textbf{Identity} & \textbf{Random} \\
\midrule
\textbf{Top} (Comp) $\uparrow$  & 0.837 & 0.862 & 4.347e-5  & 0.834 & 0.575 & 0.845 & 0.834 & 0.878  &0.959 & 5e-3 & 0.909 & 0.806 & 0.967  \\
\textbf{NT} $\downarrow$  & 2.568 & 2.703 & 8.565e-3 & 2.628 & 2.105 & 387.901 & 2.640  & 183.530 & 15.850 & 0.064 & 47.000 & 758.140 & 256.136  \\
\textbf{Ratio} $\uparrow$ & \textbf{0.326} & \textbf{0.319} & 0.051 & 0.318 & 0.273 & 0.002 & 0.316 & 0.005 & 0.061 & 0.008 & 0.019 & 0.001 & 0.003   \\
\bottomrule
\end{tabular}%
}
\caption{Softmax probability changes in downstream ICD predictions resulting from feature ablations (i.e., comprehensiveness). ``Top'' represents the mean magnitude of softmax drops for the most probable ICD code, while ``NT'' signifies the sum of absolute softmax probability changes of other ICD codes for each clinical note. The ``Ratio'' indicates the ratio between these two measures. We bold and distinguish the results obtained using our combined LAAT and dictionary learning framework, and observe that our method has the most precise effect on downstream ICD predictions, suggesting improved explanatory power.
}
\label{tab:ablation}
\end{table*}

\textbf{Addressing Superposition.} We decompose token embeddings to understand why label attention highlights specific tokens for ICD codes. But each token can hold multiple meanings, making manual evaluation of its dictionary features laborious. So, we introduce a new metric to assess if our dictionaries can identify hidden meanings even when attention identifies "extraneous" words.

\textit{\textbf{Setup.}} We analyze stop words highlighted by label attention to evaluate our dictionary's ability to explain their relevance to ICD code predictions. From 8,000 test set clinical notes, we extract \textit{all, not just attention-highlighted tokens} and build a dictionary linking tokens and ICD codes to dictionary features (described in Section \ref{sec:Method Exp}). We then focus on stop words deemed highly relevant by the label attention mechanism. For each stop word, we query its relevant dictionary features using our trained sparse autoencoder, and see if the original label ranks among the top 10 classes pre-mapped by the sampled dictionary for each activated feature. Since feature activation magnitudes vary, we define "highly activated" features as those exceeding the 96.5th percentile feature magnitude per token embedding. Additional details are listed in Appendix \ref{appendix:stopwordexp}.

\textit{\textbf{Results.}} Table \ref{table:StopWordExperiments} presents the proportion of stop word embedding labels correctly identified by our DL framework, as well as the performance of baseline methods. Notably, our DL framework, particularly the L1 sparse autoencoder variant, achieves an impressive accuracy of 91\%, outperforming all baselines. These results underscore the robustness of our DL approach in capturing the hidden medically relevant meanings embedded within superposition in stop words, which are often overlooked in traditional interpretability analyses.

\begin{table}[h]
\centering
\resizebox{0.35\textwidth}{!}{%
\begin{tabular}{@{}lc|cccc@{}} 
\toprule
\multicolumn{6}{c}{\textbf{Hidden Medical Meaning Identification Accuracy}} \\ 
\midrule
\multicolumn{2}{c|}{\textbf{ AutoCodeDL }} & \multicolumn{4}{c}{\textbf{ Baselines}} \\
\midrule
\textbf{L1} & \textbf{SPINE} & \textbf{ICA} & \textbf{PCA} & \textbf{Identity} & \textbf{Random} \\
\midrule
\textbf{0.91} & \textbf{0.89} & 0.40 & 0.53 & 0.61 & 0.37 \\
\bottomrule
\end{tabular}
}
\caption{Proportion of stop word embedding labels correctly identified by our AutoCodeDL framework, alongside the baseline methods. Such results showcase that DL is capable of effectively identifying hidden meanings embedded within superposition.
}
\label{table:StopWordExperiments}
\end{table}

\textbf{Model Steering.} Meaningful dictionary features should effectively steer model behavior by increasing the likelihood of related codes \cite{templeton2024scaling_mono}. We confirm this in the multilabel setting by "clamping" relevant feature activations, demonstrating that we can drive our model to predict specific subsets of medical codes without additional tokens, modifying model weights, or explicitly training steering vectors \cite{subramani2022steering_vectors}. This finding may inspire cheaper alternatives for quickly modifying model behavior, especially as ICD coding models become larger.
\begin{table}[h]
\centering
\resizebox{0.4\textwidth}{!}{%
\begin{tabular}{@{}l|cc|cccc@{}}
\toprule
\multicolumn{7}{c}{\textbf{Model Steering Experiment with Dictionary Features}} \\
\midrule
\textbf{Metrics} & \multicolumn{2}{c|}{\textbf{ AutoCodeDL }}  & \multicolumn{4}{c}{\textbf{Baselines}} \\
\midrule
\textbf{} & \textbf{L1} & \textbf{SPINE} & \textbf{ICA} & \textbf{PCA} & \textbf{Identity} & \textbf{Random} \\
\midrule
No. Code Flips $\uparrow$ & 3449 & 3681 & 0 & 511 & 1 & 3681 \\
No. Meaningful $f_i$ $\uparrow$ & 928 & 1353 & 0 & 10 & 1 & 768 \\
ID Accuracy $\uparrow$ & \textbf{0.55} & \textbf{0.89} & 0.29 & 0.32 & 0.26 & 0.26 \\
\bottomrule
\end{tabular}
}
\caption{Table \ref{table: Model Steering}: Model steering experiment results comparing AutoCodeDL (L1 and SPINE) with baselines. Effectively changing all medical codes and still attaining high identification accuracy through the discovery of clamped classes, AutoCodeDL with SPINE is capable of steering the model in highly interpretable ways.}
\label{table: Model Steering}
\end{table}

\textit{\textbf{Setup.}} To measure the direct impact of each dictionary feature on ICD code predictions, we input pad tokens to generate a blank canvas of PLM embeddings. Instead of ablating decomposed dictionary features (eq. \ref{eq:feature ablation}), we manually "clamp" or increase each feature's activation to a large value (50) and reconstruct new embeddings. We then measure the increases in downstream ICD code probabilities, counting the number of ICD codes with a softmax probability increase of 0.5 or more (i.e., a decision flip) and their respective number of dictionary features. To validate the meaningfulness of each clamped dictionary feature, we construct a new dictionary with the top ICD code probabilities increased by each clamped feature and rerun the hidden meaning identification experiment from Table \ref{table:StopWordExperiments}.

\begin{figure*}[ht]
\centering
\includegraphics[width=1.0\textwidth]{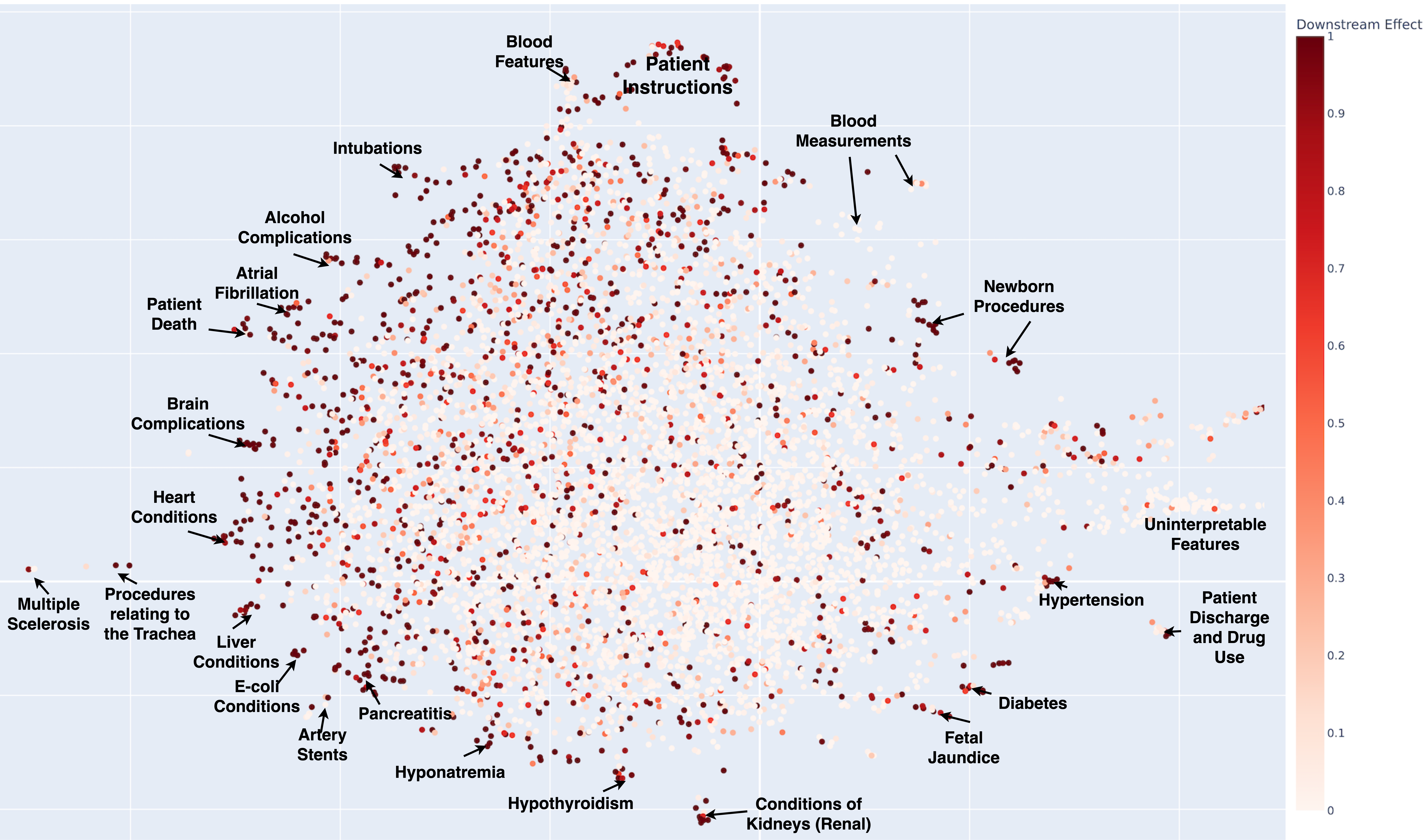}
\caption{UMAP of SPINE Embeddings: Dictionary features are interpretable in steering model behavior. More darker red colors indicates higher maximum observed probability increase of the top medical code from a feature's exclusive clamping. Each dot is a dictionary feature embedding projected into 2D. }
\label{fig:umap}
\end{figure*}

\textit{\textbf{Results.}} Table \ref{table: Model Steering} shows that while not all dictionary features are meaningful in steering model behavior, the dictionary embeddings can change the predictions of nearly all ICD codes. Moreover, the clamped classes used to generate the dictionary can still recover the hidden medical codes of extraneous stop tokens, suggesting each feature's explainability. Further verification is provided by a UMAP plot (Figure \ref{fig:umap}), where the color indicates the maximum increase in probability of its top medical code. The plot reveals clusters of semantically meaningful features with a direct ability to change model predictions of specific subsets of codes, annotated based on their dictionary contexts and relevant medical codes.

\vspace{-0.1cm} 
\subsection{Human Understandability of Dictionaries} \label{sec:h-understandable}
Evaluating the human understandability of interpretability methods lacks a clear definition, often depending on qualitative assessments \cite{Zhang_2021SurveyNNInterpretability, info14080469SurveyEvaluationInterpretabilityMethods}. Given the limited availability of clinically licensed physicians, there's a need for scalable proxy metrics to measure the understandability of learned dictionaries across extensive clinical notes, encompassing thousands of tokens. Through manual examination of various dictionary features, we pinpoint two primary factors that contribute to a dictionary feature's understandability.

\textbf{Coherence.} A dictionary feature is deemed understandable if its top tokens are semantically related, indicating a clear conceptual identity. Conversely, semantic randomness among tokens complicates the identification of a feature's underlying concept. 

\textit{\textbf{Setup.}} Advancements in Siamese encoder transformers, like Siamese BERT, have boosted the efficiency and effectiveness of semantic similarity analyses \cite{reimers2019sentencebert_siamese_bert}, mimicking human perceptions of text pair similarity. Utilizing Siamese encoder embeddings, we calculate the average cosine similarity among the top $k$ tokens of each dictionary feature to gauge their conceptual relatedness. For more methodological specifics of our Siamese BERT experiment, please refer to Appendix~\ref{appendix:siamese_cosine}. Although LAAT is part of our method for better explaining downstream ICD predictions, we also treat the LAAT matrix as a "dictionary" of ICD codes, similar to dictionary features learned by sparse autoencoders. While comparing supervised (LAAT) and unsupervised (autoencoders) methods isn't perfect, LAAT remains the current standard for coherence in this area of automated ICD coding with PLMs.

 \textit{\textbf{Results.}} The coherence of the unsupervised dictionary learning (DL) methods, represented by the DL columns, decreased as the number of top k tokens increased. Contrary to expectations from \citet{subramanian2017spine}, sparse autoencoders with L1 minimization exhibited the highest coherence among unsupervised methods, with the highest average cosine similarity across different k values. Qualitative examples and the relationship between dictionary contexts and coherence are provided in Appendix \ref{appendix:cosine_qualitative_human_eval}. The most interpretable dictionary features correspond to high cosine similarities, potentially filtering out easy cases for human annotators to focus on more complex features.

\begin{table}[h!]
\centering
\resizebox{0.5\textwidth}{!}{%
\begin{tabular}{@{}lcc|cccc|cc@{}}
\toprule
\multicolumn{8}{c}{\textbf{Coherence of All Activated Dictionary Features}} \\ 
\midrule
\multicolumn{1}{c}{}  & \multicolumn{2}{c|}{\textbf{DL}} & \multicolumn{4}{c|}{\textbf{Baselines}} & \multicolumn{1}{c}{\textbf{Supervised}} \\
\midrule
$k \uparrow$ &\textbf{L1} & \textbf{SPINE} & \textbf{ICA} &\textbf{PCA} & \textbf{Identity} &\textbf{Random} & \textbf{LAAT} \\
\midrule
\textbf{2} & \textbf{0.3130} & 0.3074 & 0.2138 & 0.2371 & 0.2498 & 0.2591 & 0.4185 \\
\textbf{4} & 0.2981 & 0.2872 & 0.2133 & 0.2294 & 0.2440 & 0.2449 & 0.4083 \\
\textbf{10} & 0.2747 & 0.2684 & 0.2095 & 0.2218 & 0.2384 & 0.2358 & 0.3889 \\
\bottomrule
\end{tabular}
}
\caption{Average cosine similarity between the top k tokens extracted from each dictionary feature or ICD code, measured from Siamese encoder embeddings. Higher values indicate a stronger thematic connection within the feature or code. The "DL" columns represent our dictionaries constructed, while the remaining columns are baselines.}
\label{table:tokens_compared}
\end{table}

\textbf{Distinctiveness.} If a dictionary feature has a clear, coherent theme based on its top tokens, unrelated tokens should be readily discernible.

\textit{\textbf{Setup.}} Inspired by \citet{subramanian2017spine}, who evaluated the distinctiveness of dictionary features in sparse word embeddings, we investigate by sampling a dictionary feature's top 4 ($k$=4) activating tokens (and their nearby context windows) and a randomly sampled token outside the feature. An interpretability score is derived by having medical experts (a licensed physician and a medical scientist trainee) identify the randomly sampled token from the set of 5 tokens. The proportion of correctly identified tokens serves as our distinctiveness metric. Due to time constraints, our medical experts evaluated 100 samples each for DL and other baseline encoders. However, given the demonstrated capabilities of state-of-the-art language models in text annotation tasks \cite{Huang_2023_CHatGPTHumaAnnotators, Gilardi_2023_outperforms_text_annotation} and their extensive medical vocabulary \cite{Bommineni2023.03.05.23286533_MCAT_CHATGPT}, we utilized the current medically quantized state-of-the-art OpenBioLLM Llama 3 70B model across all dictionary features from the dictionaries constructed in the stop words experiment (Section \ref{sec:Method Exp}).

\textit{\textbf{Results.}} As shown in Table \ref{table:GPT3.5exp}, both sparse autoencoders are more distinctive than their respective unsupervised baselines. Surprisingly, the random and identity encoders are more distinguishable than their PCA counterparts. We provide examples in Appendix ~\ref{appendix: human evaluations distinctiveness} illustrating that the ability to differentiate features can range from exceptionally obvious cases with repeating tokens or differences in specificity to completely uninterpretable cases.

\begin{table}[h!]
\centering
\resizebox{0.5\textwidth}{!}{%
\begin{tabular}{l|cc|cccc|c}
\toprule
\multicolumn{8}{c}{\textbf{Percentage of Dictionary Features Differentiated }} \\
\midrule
\multicolumn{1}{c}{}  & \multicolumn{2}{c|}{\textbf{DL}} & \multicolumn{4}{c|}{\textbf{Baselines}} & \multicolumn{1}{c}{\textbf{Supervised}} \\
\midrule
& \textbf{L1} & \textbf{SPINE} & \textbf{ICA} & \textbf{PCA} & \textbf{Identity} & \textbf{Random} & \textbf{LAAT} \\
\midrule
No. LLM Id. $\uparrow$ & 2828 & 2713 & 282  & 226  & 297 & 299 &  2117 \\
\% LLM Id. $\uparrow$ & 0.46 & 0.44 & 0.37 & 0.29 & 0.39 & 0.39 & 0.58 \\
\% Human Id. (100) $\uparrow$ & 0.49 & 0.56 & 0.45* & 0.41 & 0.45 & 0.44 &  *\\
\bottomrule
\end{tabular}
}
\caption{Percentage of dictionary features successfully distinguished by the quantized OpenBioLLM-70B Llama 3 model and medical experts, determined by correctly identifying the unrelated token from a set of 5 tokens (including the top 4 activating tokens and their context windows) highlighted by our sparse autoencoders. * denotes cases where human evaluations were omitted or reduced (i.e., 40) due to time constraints.}
\label{table:GPT3.5exp}
\end{table}


%% file: sections/conclusion.tex
\vspace{-0.2cm} 
\section{Conclusion} \label{sec:Conclusion}
This study introduces a novel method that combines dictionary learning with label attention mechanisms to improve the interpretability of medical coding language models. By uncovering interpretable dictionary features from dense language embeddings, the proposed approach offers a dictionary-based rationale for ICD code predictions, addressing the growing demand for transparency in automated healthcare decisions. This work lays the groundwork for future explorations in dictionary learning to enhance healthcare interpretability.

%% file: sections/appendix.tex
\section{Appendix} \label{sec:Appendix}
\subsection{Sparse Autoencoder Training Details} \label{appendix:saenc_training}
We train our sparse autoencoders on the PLM activations generated by a 110M medical RoBERTa encoder PLM on the cleaned MIMIC-III train dataset of 38,427 clinical notes and evaluated on their test set of 8,750 clinical notes for a total of 52,712 clinical notes, as detailed by \cite{AutomatedMedicalCoding}. We follow the advice of \cite{bricken2023monosemanticity} and \cite{subramanian2017spine} in training the $L_1$ and SPINE autoencoders respectively. Our hyperparameters are shown in Table \ref{tab:training detail}. We also use AdamW as our optimizer. For a fair comparison, we reuse the dictionary feature size $m$ as it has been noted by \cite{bricken2023monosemanticity} that larger dictionary feature sizes can potentially increase the resolution of concepts of dictionary features. For instance, dictionary feature may be further decomposed into features of of more specific meanings given different token sequence contexts as further discussed by \cite{bricken2023monosemanticity}. We use PyTorch as our deep learning framework of choice.

\begin{table}[ht] 
\centering
\caption{Sparse Autoencoder Training Details} \label{tab:saenc_training}
\label{tab:training detail}
\label{your-table-label}
\begin{tabular}{@{}cccccc@{}} 
\toprule
$\lambda_{L_1}$ & $\lambda_1$ & $\lambda_2$ & $m$ & Batch Size & $lr$ \\
\midrule
2e-5 & 1 & 1 & 6,144 &  8,192 & 1e-3 \\
\bottomrule
\end{tabular}
\end{table}

We train on randomly sampled embeddings from the training set, and filter out all pad tokens, that are irrelevant to the final prediction, but dominate each batch due to their use in making GPU inference fast. We note that we cannot get perfect reconstruction loss nor completely match the downstream performance of the original model with our reconstructed embeddings on the test set, but we get very close. 

\begin{table}[ht]
\centering
\caption{Autoencoder Test Loss Metrics}
\label{table:performance_metrics}
\resizebox{0.5\textwidth}{!}{%
\begin{tabular}{lccc} 
\toprule
& L1 & SPINE & Original\\ 
\midrule
Test Autoencoder Loss & 69.46 &  39.59 & \textbf{N/A} \\ 
Test F1 & 0.258 & 0.260 & 0.262 \\ 
\bottomrule
\end{tabular}
}
\end{table}

\subsection{Scalability Discussion}
Training these sparse autoencoders takes approximately 6 minutes per epoch on A6000 GPUs. However, the current training process samples new token embeddings for every batch of clinical notes, which is suboptimal. A quick adaptation of existing dataloaders to improve token diversity during training could be beneficial. We find that precomputing and caching PLM embeddings can significantly reduce training time to approximately 15 minutes for 10 epochs, albeit requiring substantial memory (at least 128 GB RAM for caching millions of embeddings). In contrast, decomposing PLM embeddings using our method is extremely fast, on par with LAAT (approximately 0.04 seconds per clinical note). While there is an upfront cost (a couple of hours for sorting and sampling millions of tokens for each dictionary feature), once a dictionary is constructed, interpreting the embedding space of any clinical note is very fast and efficient, which is suitable for this high dimensional multilabel task.

\subsection{Build Dictionary} \label{appendix:build_dict}
For further clarity, we write up our dictionary construction algorithm here in algorithm \ref{alg:build_dict}. In principle, one is just sorting based on encoded activations and ablation softmax drops of different ICD codes for each dictionary feature.
\begin{algorithm}[h!]
\caption{Build Dictionary}
\label{alg:build_dict}
\KwIn{Autoencoder $A$, feature $f_i$, tokens $x$}
\KwOut{Dictionary $F$ mapping $f_i$ to tokens $x$ and classes $y$}

$F \gets \text{dict}$;
\For{each token $x$}{
$f \gets A.encode(x)$;
\For{$f_i$ in $f$}{
\If{$f_i > F[i].f_i$}{
$F[i].tokens \gets x$;
}
$\delta_i \gets \text{ablation}(f_i)$;
\If{$\delta_i > F[i].\delta$}{
$F[i].classes \gets \text{drops}(\delta_i)$;
}
}
}
\Return $F$;
\end{algorithm}

\subsection{Additional Baseline Details} \label{appendix:baselines}
There were four main baseline methods that were compared against our exploration of two sparse autoencoders, specifically an ICA encoder, PCA encoder, an identity encoder, and a random encoder. All of their implementations were taken from \cite{cunningham2023sparse}. The ICA encoder was trained using the FastICA decomposition method from scikit-learn to estimate the activation's respective independent components that act as the encoder weights. However, we note that the training was unstable, most likely due to the same memory and computation limitations that \cite{cunningham2023sparse} faced. As a result, we limit the training on only 2,000,000 token embeddings sampled from the training set. The PCA covariance matrix was estimated batchwise with its eigenvectors acting as the encoder weights. For the random and identity encoders, the random encoder was initialized with a normal distribution of mean 0 and variance 1 and the identity encoder has an identity matrix for its encoder weights. 

\subsection{Label Attention Details} \label{appendix:laat}
\begin{figure*}[ht]
\centering
\includegraphics[width=1.0\textwidth]{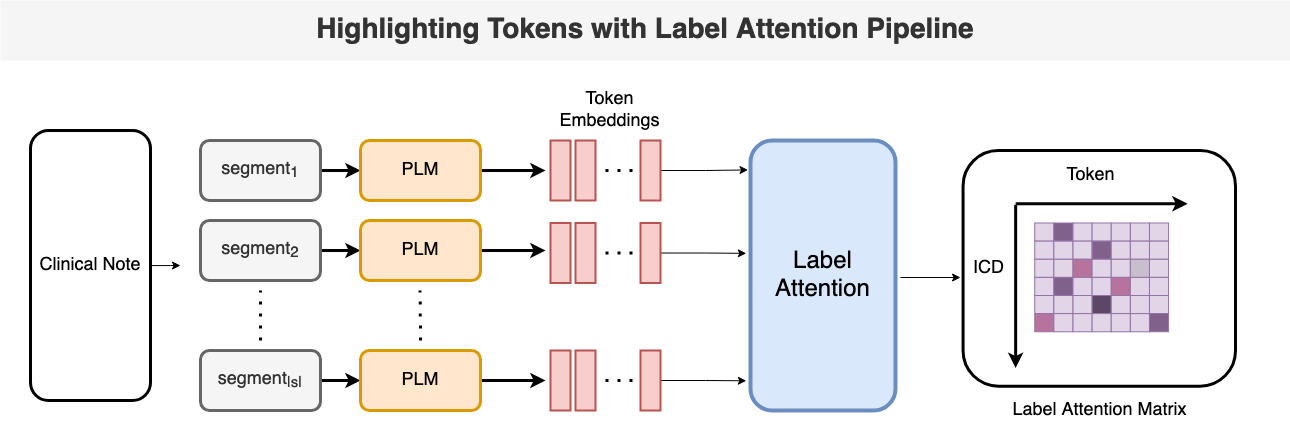}
\caption{Label Attention identifies the most relevant tokens for each ICD code through a label attention matrix. }
\label{fig:laat}
\end{figure*}

We recognize that we don't explicitly describe LAAT \cite{Vu_2020_LAAT, huang2022plmicd} in detail in the main manuscript. For interested readers, we depict the label attention mechanism in Figure \ref{fig:laat}.

Essentially, LAAT computes a cross attention score for each token and ICD code, creating a label attention matrix where each row is an ICD and every column is the token's attention score with respect to that ICD code. 

\subsection{Additional SPINE Details} \label{appendix:spine}
While the $L_1$ minimization is most commonly used to train sparse autoencoders due to its simplicity of training \cite{Zhang_2015_Survey_Sparse_Representations}, we also revisit an alternative sparse formulation SPINE proposed by \cite{subramanian2017spine}, specifically designed to decompose neural word embeddings. \cite{subramanian2017spine} showed that they could improve interpretability in GloVe \cite{pennington-etal-2014-glove} and other forms of neural token embeddings \cite{almeida2023word-embeddings}. Their formulation replaces the $L_1$ regularization loss in favor of using a combination of average sparsity and a partial sparsity loss function terms. We outline their method in the equations below.

We define our average sparsity term below in equation \ref{eq: ASL}.
\begin{equation} \label{eq: ASL}
    \mathcal{L}_{\text{asl}} =  \sum_i \max(f_i, \rho) 
\end{equation}
Here, $\rho$ is a user-chosen sparsity hyperparameter that prevents the dictionary feature activations $f$ from becoming too large. In practice, one should ideally average each sparse feature activation $\bar{f_i}$ across the entire training set first as outlined by \cite{subramanian2017spine}. However, due to memory constraints, we simply average over the batch size.

The partial sparsity term is defined as follows below.
\begin{equation}
    \mathcal{L_\text{psl}} = \sum_i f_i \cdot (1 - f_i)
\end{equation}
The general intuition behind the following partial sparsity term is that it further enforces sparsity by penalizing the distribution of $f_i$'s away from uniformity, which can invariably occur due to the stochastic nature of gradient optimization algorithms. The resulting alternative sparse autoencoder minimization function is shown in equation \ref{eq : spine} .

\begin{equation} \label{eq : spine}
\mathcal{L} = \frac{1}{|X|} \sum_{x \in X} \left\lVert x - \hat{x} \right\rVert_2^2 + \lambda_1  \mathcal{L}_{\text{asl}} + \lambda_2 \mathcal{L_\text{psl}}    
\end{equation}

\subsection{Baseline Ablation Experiment Details} \label{appendix:baseline ablation details}
For every baseline, we iterate through each activated dictionary feature for the tokens, measuring the softmax drop for the selected ICD code (with the highest softmax probability). This is repeated for each none-DL baseline: In ICA, we ablate embedding independent components. In PCA, we ablate principal components (i.e eigenvectors weighed by their eigenvalues). In the identity encoder, we ablate embedding dimensions. In the random encoder, we essentially add random noise. For the Token baseline, we ablate entire token embeddings. 

\subsection{Sufficiency Discussion}
If comprehensiveness measures the performance drop when ablating key features, sufficiency measures the performance retention when keeping only the key features \cite{chan-etal-2022-comparative_faithfulness}. In particular, rather than performing an ablation, we simply set all of the token embeddings to their respective dictionary feature embeddings multiplied by their activations. Clinical coding is high-dimensional, making it computationally expensive to compute different quantiles. Therefore, we only compute metrics for the highly relevant 95\% quantile tokens when using LAAT, or for all tokens when not using LAAT.
\begin{equation} \label{eq:feature sufficiency}
    \Tilde{x} = f_i \cdot h_{i}
\end{equation}
 However, when using our specific feature encoders, this metric yields mixed results. While retaining only the relevant token embeddings defined by LAAT preserves the model's class probabilities, we find that a randomly generated embedding from our random encoder remarkably performs similarly or better than the LAAT explanation. This observation raises further questions about the potential inner workings and importance of the embedding space in explaining model behavior. Especially, as the steering experiment is effectively related to the sufficiency metric, but with difference being $f_i$'s artificial expansion versus its original encoding. Understanding these challenges in embedding interpretability is an important point of future work.

\begin{table*}[h!]
\centering
\resizebox{1.0\textwidth}{!}{%
\begin{tabular}{c|c c |c c c c c | c c |c c c c}
\toprule
\multicolumn{13}{c}{\textbf{Ablating Dictionary Features of Highlighted Tokens}} \\
\midrule
\multicolumn{1}{c|}{\textbf{Experiment}} & \multicolumn{2}{c|}{\textbf{ AutoCodeDL }} & \multicolumn{5}{c|}{\textbf{LAAT + Baselines }} & \multicolumn{2}{c}{\textbf{ DL }}  & \multicolumn{4}{|c}{\textbf{ Baselines }} \\
\midrule
 & \textbf{L1} & \textbf{SPINE} & \textbf{ICA} &  \textbf{PCA} & \textbf{Identity} & \textbf{Random}  & \textbf{Token} & \textbf{L1} & \textbf{SPINE} & \textbf{ICA} &  \textbf{PCA} & \textbf{Identity} & \textbf{Random} \\
\midrule
\textbf{Suff. $\downarrow$}   & 0.143 & 0.462 & 0.470 & 0.424 & 0.468 & -0.029 & -0.012 & 0.137   & 0.454 & 0.470 & 0.112 & -0.005 & -0.033   \\
\bottomrule
\end{tabular}%
}
\caption{Softmax probability changes in downstream ICD predictions resulting from sufficiency experiments.}
\label{tab:ablation_old_gt_sufficiency}
\end{table*}

\subsection{Hidden Meaning Stop Words Experiment} \label{appendix:stopwordexp}
We explain more details of our stop words experiments here. To begin, we sample \textit{all (not just the attention highlighted)} tokens from 1,600 clinical notes sampled from the test set and then map highly activating tokens to each dictionary feature as well as relevant ICD code predictions through dictionary feature ablations outlined in section \ref{sec:Method Exp} and \ref{sec:BuildingDictionary}. Then, we collect (12,891) highly relevant stop words  (using NLTK) identified by the label attention mechanism, documenting their corresponding labels and PLM embeddings as defined by the label attention matrix. After shuffling each label-embedding pairing, we employ our dictionary to query the stop word's relevant classes via a trained sparse autoencoder, assessing if the original label ranks among the top 10 classes of each of its most highly activated dictionary features. As the magnitudes of different dictionary features varies across decomposition methods and are mostly sparse in sparse autoencoders, we define highly activated features $f_i$ as dictionary features that exceed the 96.5th percentile feature magnitude of encoded from each token embedding. In practice, since the encoded dictionary features for each token by sparse autoencoders are sparse, the highest activated dictionary features are those that are nonzero. The efficacy of dictionary features in clarifying a stop word's significance to an ICD code prediction is quantified by calculating the proportion of shuffled stop word embeddings are correctly contained in these prediction sets generated by their respective dictionaries. 

\subsection{Additional Siamese BERT Cosine Similarity Experiment Details} \label{appendix:siamese_cosine}
We use "all-mpnet-base-v2" from the sentence-transformers package as our Siamese encoder. We show our steps for computing the cosine similarity scores in our evaluation of coherence scores below in algorithm \ref{alg:Siamese}. Furthermore, since sparse encoding is a more efficient operation than searching and ablating the most relevant dictionary feature for an ICD prediction, we sample all tokens from every clinical note in the test set, and discern all of the top tokens for each dictionary feature with our sparse autoencoder.

\begin{algorithm} [h!]
\caption{Cosine Similarity Score for Dictionary Features}
\label{alg:Siamese}
\KwIn{Siamese BERT model $M$, dictionary feature $f_i$\;  $k$ number of highly activating tokens in dictionary $T = \{t_{f_i,1}, t_{f_i,2}, \dots t_{f_i,k}\}$ \;}
\KwOut{Cosine Similarity Score $\bar{s}$}
$\bar{s} \gets 0 $ \;
\For{each $f_i$}{
    $ t_\text{pairs} \gets \{ (t_{f_i,a}, t_{f_i,b}) \mid a \neq b, t_{f_i, a}, t_{f_i,b} \in T \} $)\;
    \For{each $(t_{f_i,a}, t_{f_i,b} )$ in $t_{\text{pairs}}$}{
        $\hat{s} = \frac{M(t_{f_i,a}) \cdot M(t_{f_i,b}) }{\|M(t_{f_i,a})\| \cdot \| M(t_{f_i,b})\|}$ \;
        $\bar{s} \gets \bar{s} + \frac{\hat{s}}{|t_{pairs}|}$ \;
    }
    $\bar{s} \gets \frac{\bar{s}}{|f|}$
}
\end{algorithm}

\subsection{Human Evaluations} \label{appendix: Human Eval}
We conducted human evaluations with a medical scientist trainee and a licensed physician, specifically the distinctiveness experiment inspired by \citet{subramanian2017spine}. Below, we showcase examples of incoherent dictionary features (low cosine similarity) and highly coherent features (high cosine similarity). We also provide examples from the human evaluations, including cases where annotators failed to identify the randomly chosen context and cases where it was easy for them to distinguish the random token.

\newpage
\subsubsection{Dictionary Features and Coherence } \label{appendix:cosine_qualitative_human_eval}
\begin{figure}[H]
\centering
\includegraphics[width=0.45\textwidth]{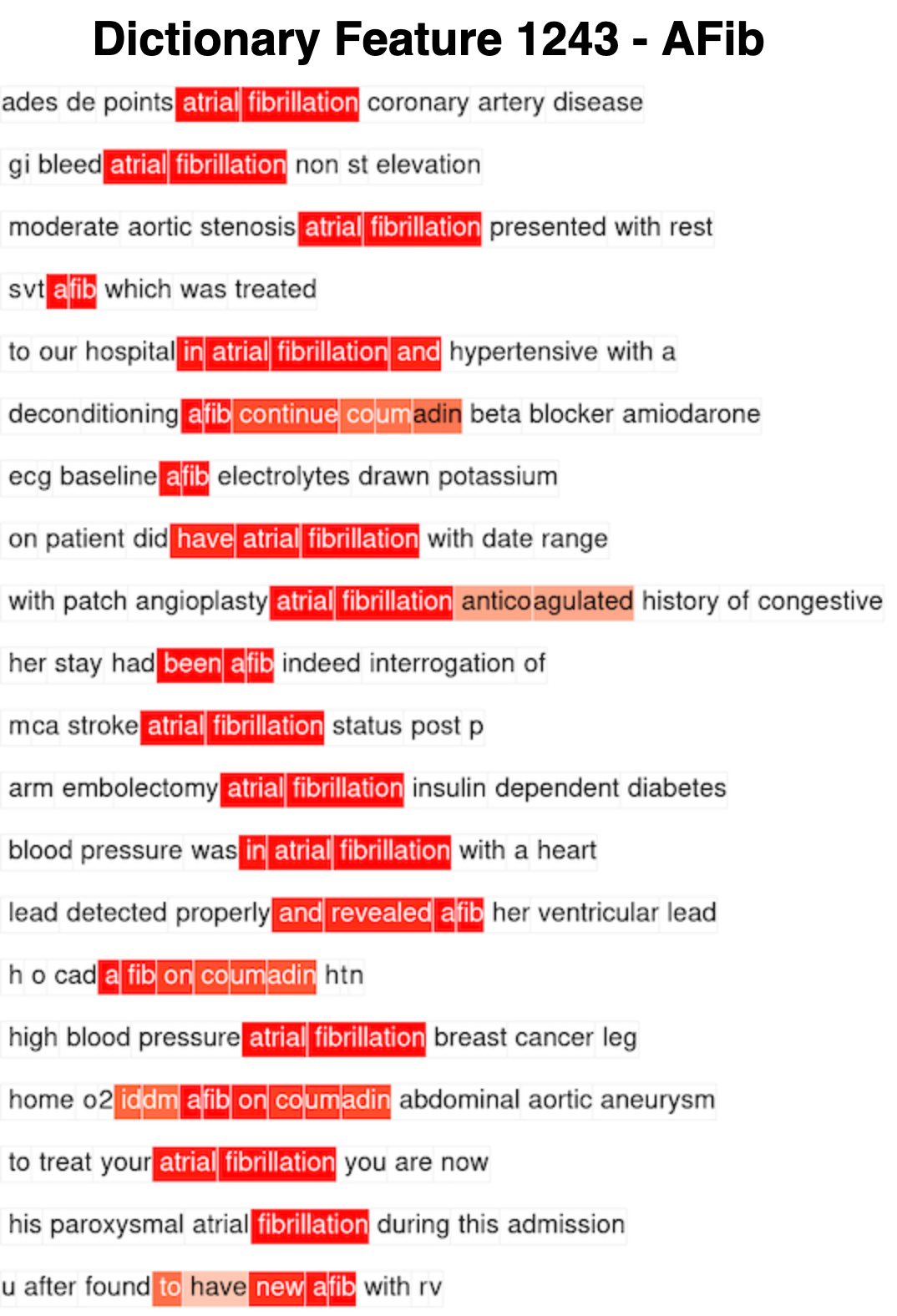}
\caption{\textbf{Example of highly interpretable SPINE feature with high cosine similarity:} In this particular case, all activating tokens (\textcolor{red}{red}) with their context windows are all atrial fibrillation tokens, giving us a very high cosine similarity (i.e. close to 1).
}
\label{fig:ez_dict_feat}
\end{figure}

\begin{figure}[H]
\centering
\includegraphics[width=0.45\textwidth]{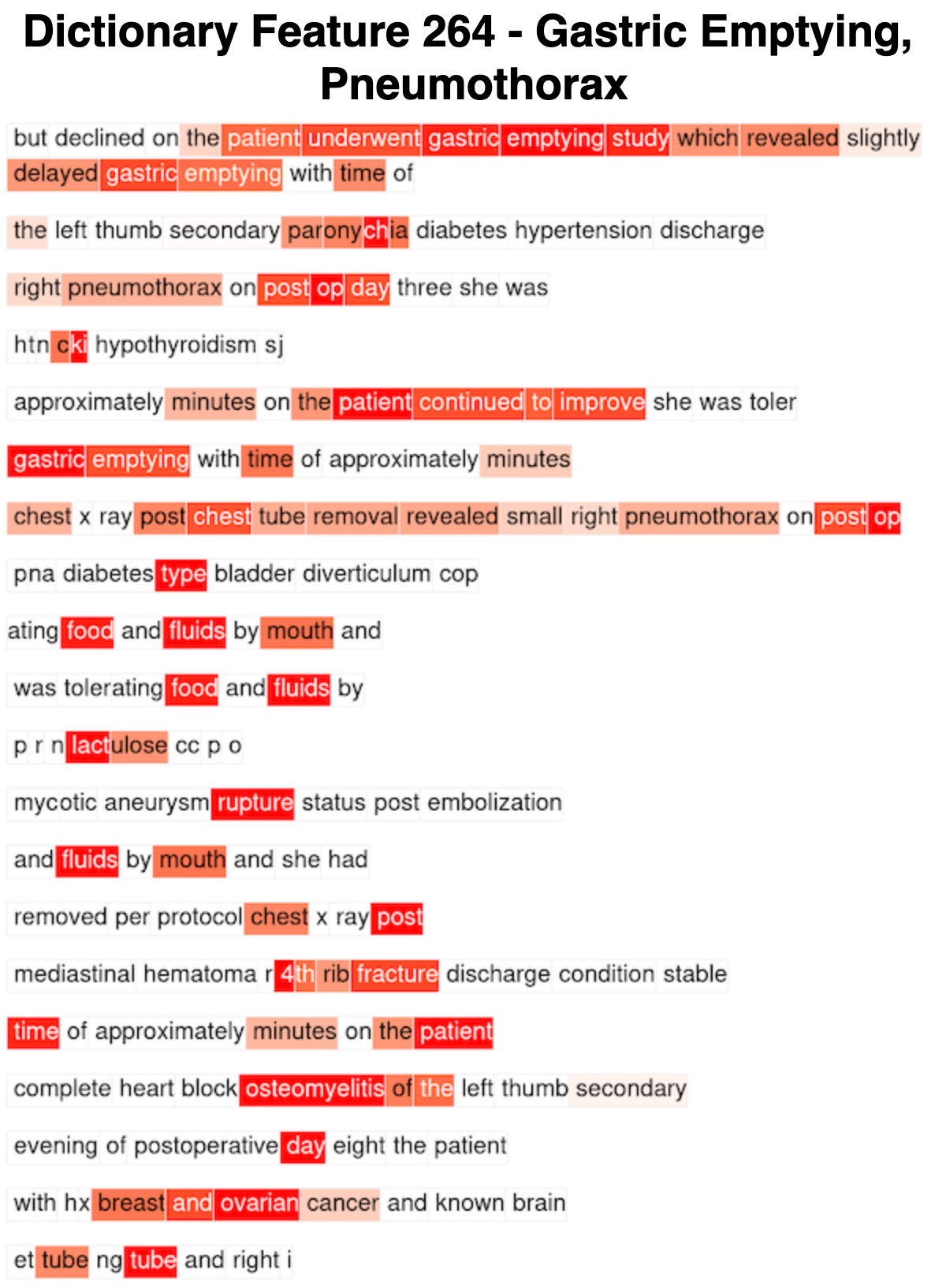}
\caption{\textbf{Example of interpretable SPINE feature with low cosine similarity:} In this case, the activating tokens (\textcolor{red}{red}) are diverse, containing various concepts such as food, fluids, gastric emptying, pneumothorax, and related terms. Our clamping experiments show that this dictionary feature is predictive of the code "acidosis," a common complication potentially leading to gastric emptying. For such cases, the cosine similarity metric is not informative, as these coherent features often have lower cosine similarities (closer to 0) despite sharing a common theme.}
\label{fig:mid_dict_feat}
\end{figure}

\begin{figure}[H]
\centering
\includegraphics[width=0.45\textwidth]{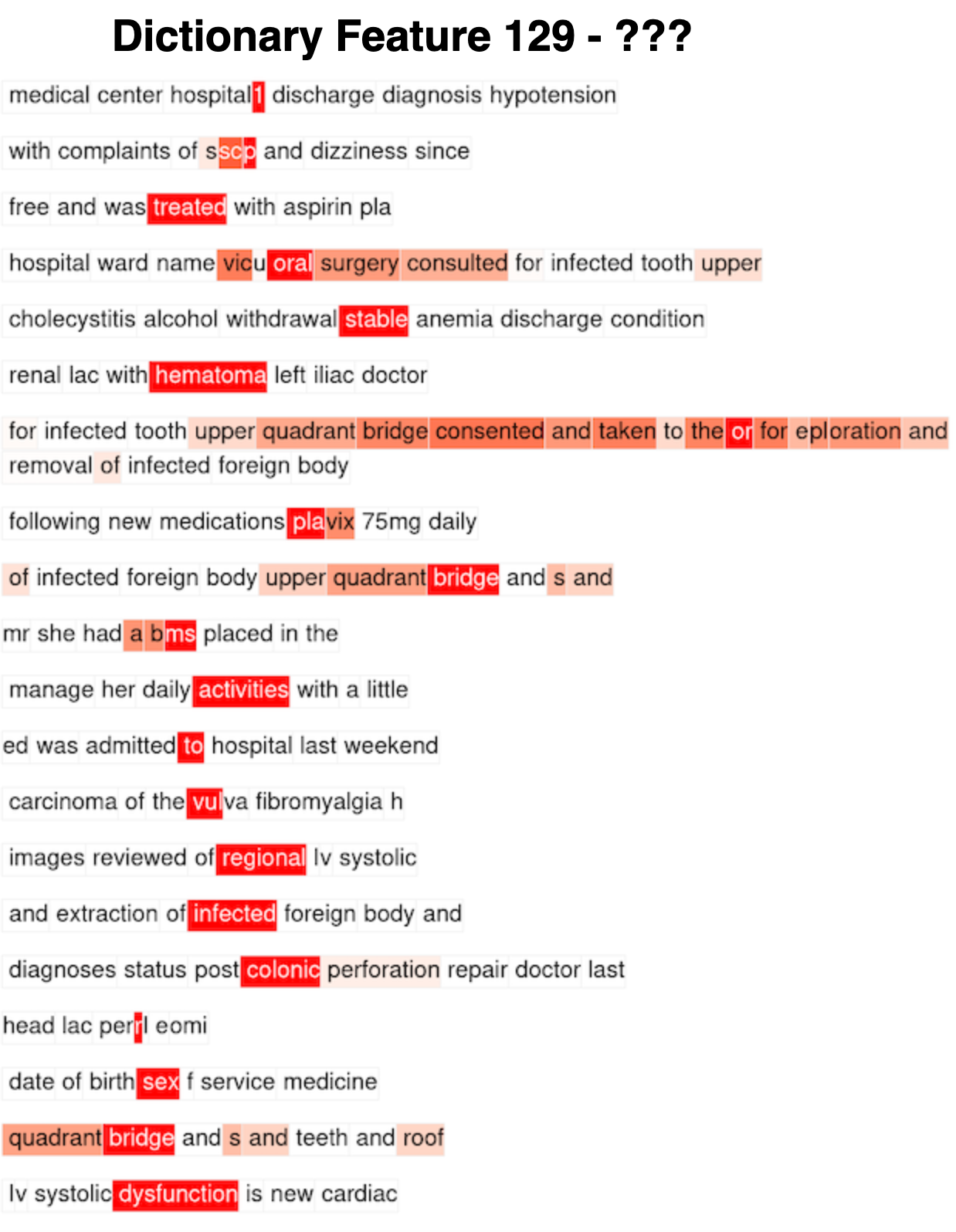}
\caption{\textbf{Example of an uninterpretable SPINE feature with low cosine similarity:} Various tokens are highlighted (\textcolor{red}{red}) without an obvious cohesive theme. As expected, these tokens result in very small cosine similarity measurements.}
\label{fig:hard_dict_feat}
\end{figure}

\vspace{-0.2cm} 
We provide qualitative examples in Figures \ref{fig:ez_dict_feat}, \ref{fig:mid_dict_feat}, and \ref{fig:hard_dict_feat} to demonstrate the efficacy of our cosine similarity metric. While an imperfect metric, cosine similarity can effectively discern highly interpretable dictionary features, which typically have repeating tokens. However, many dictionary features are more challenging and may require more complex annotation approaches, such as domain-specific language models, medical experts, or fine-tuning specific Siamese BERT encoders for this task, as the average cosine similarity of the top k tokens is intrinsically low due to the diversity of the text.

\newpage
\subsubsection{Human Distinctiveness Case Studies} \label{appendix: human evaluations distinctiveness}
\begin{figure}[H]
\centering
\includegraphics[width=0.45\textwidth]{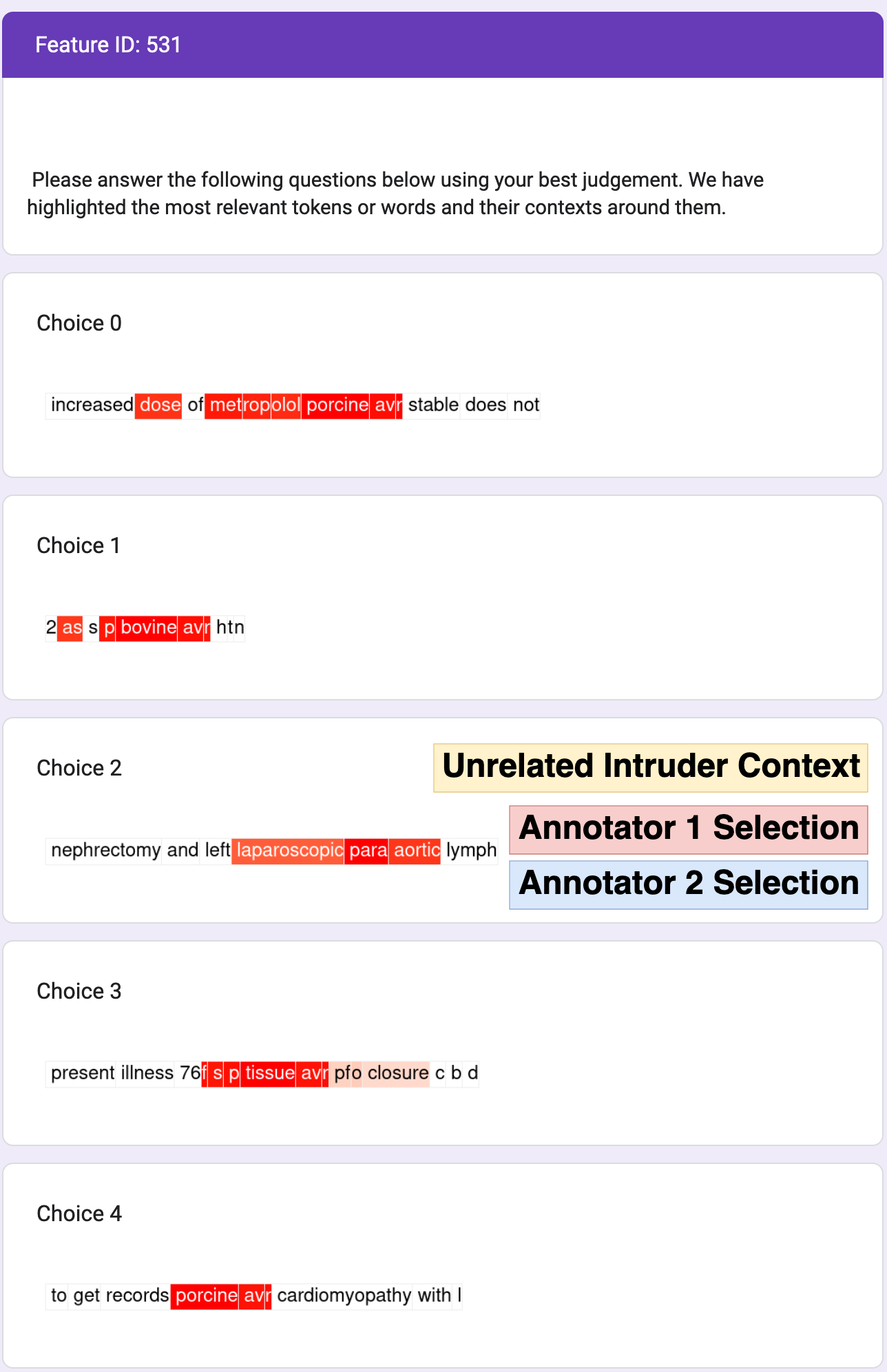}
\caption{\textbf{Example of an interpretable $L_1$ feature.} All highly activating tokens are related to heart conditions whereas the token "laparoscopic", an operation performed on the abdomen was clearly an outlier. We also observe the repeating abbreviations "avr". }
\label{fig:ez_l1_dict_feat}
\end{figure}

\begin{figure}[H]
\centering
\includegraphics[width=0.45\textwidth]{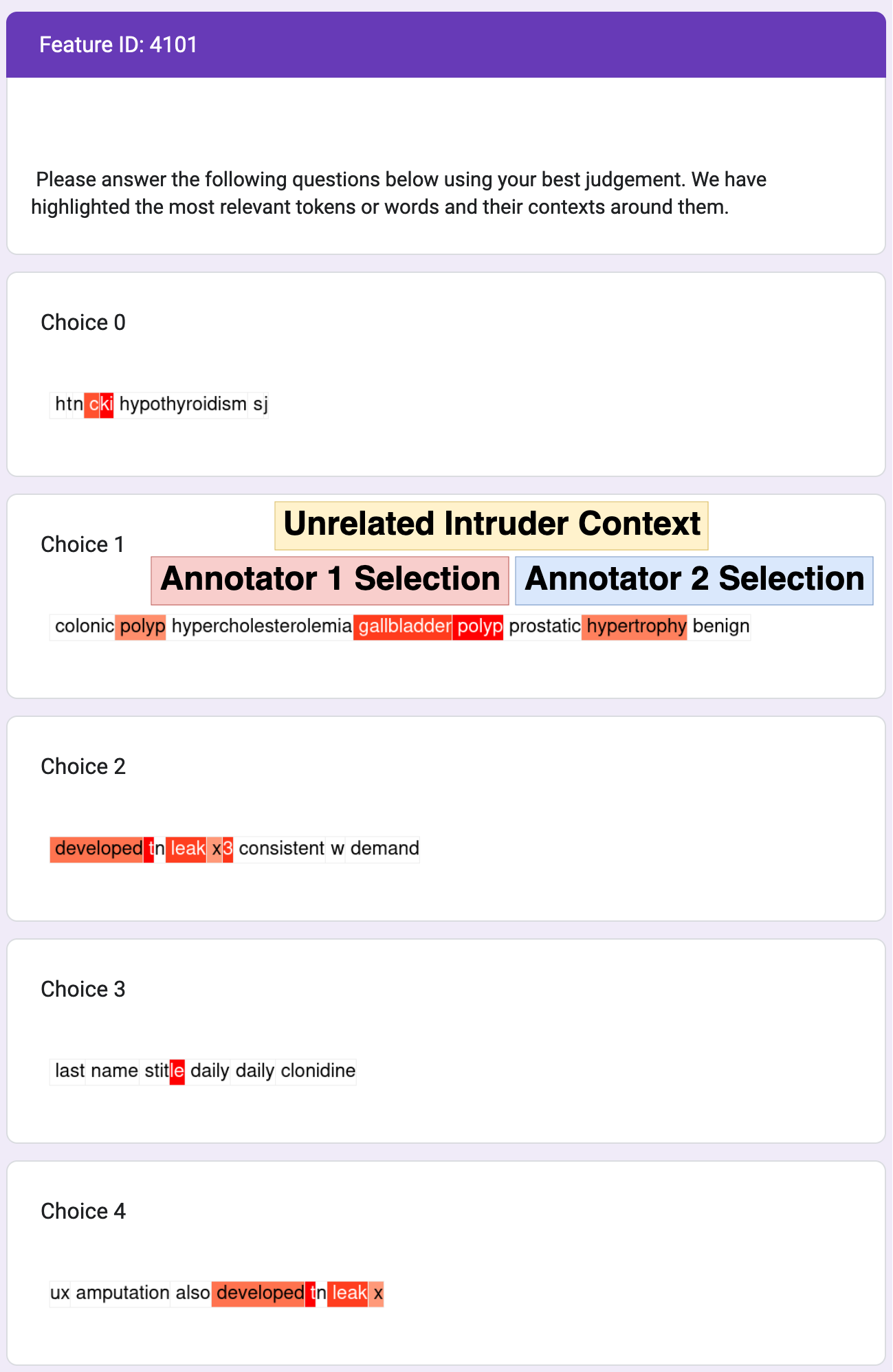}
\caption{\textbf{Example of a less interpretable $L_1$ feature differentiated by both annotators:} While both annotators acknowledged the lack of an explicit medical theme, they selected the set of tokens discussing a specific piece of anatomy, suggesting that the feature's interpretation activates a more abstract concept.}
\label{fig:med_l1_dict_feat}
\end{figure}

\begin{figure}[H]
\centering
\includegraphics[width=0.45\textwidth]{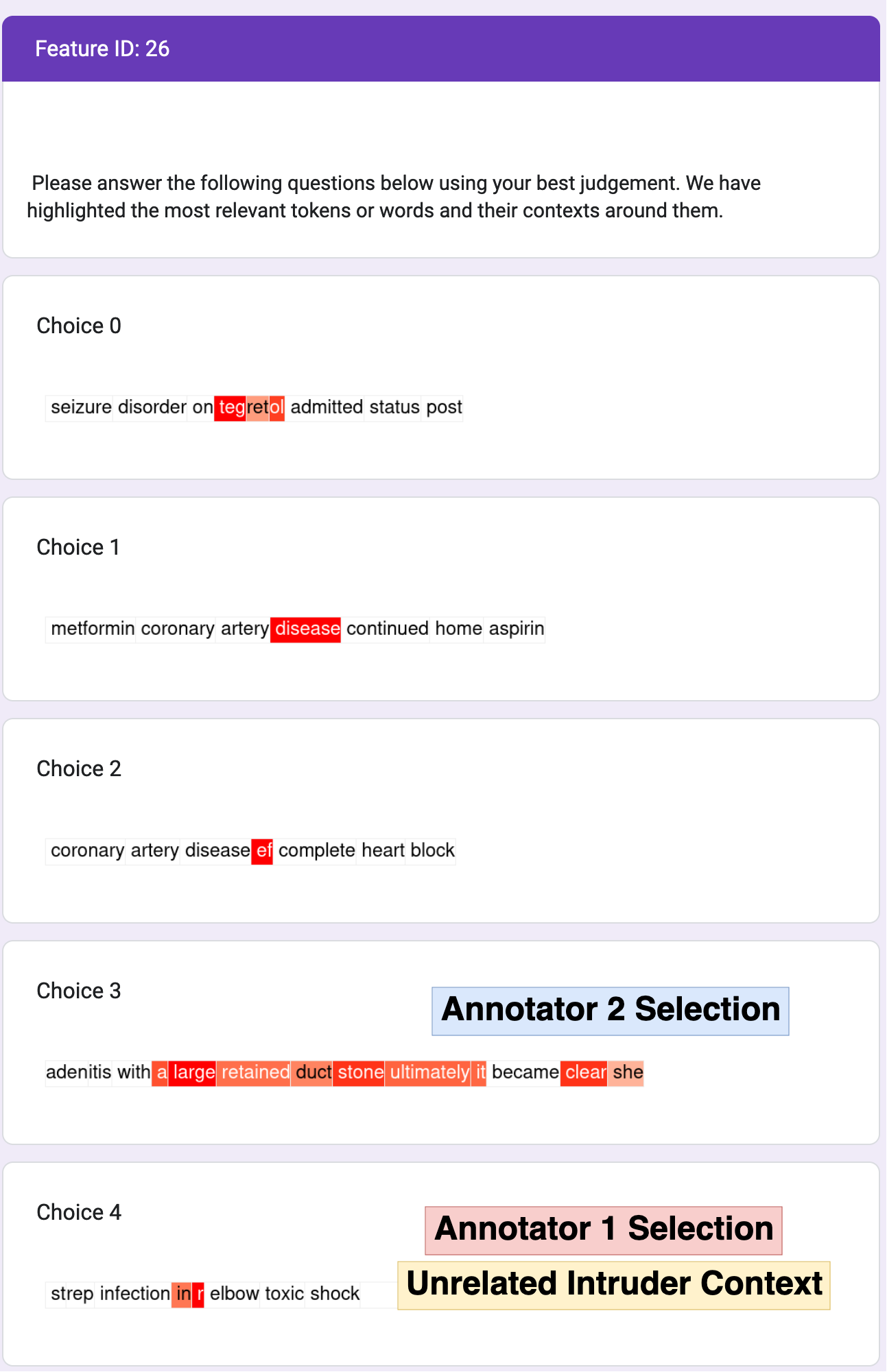}
\caption{\textbf{Example of an uninterpretable $L_1$ feature where only one annotator identified the random context:} According to our experts, in such cases where no discernible underlying theme exists, selecting the random context is essentially by chance.}
\label{fig:med_l1_dict_feat2}
\end{figure}

\begin{figure}[H]
\centering
\includegraphics[width=0.45\textwidth]{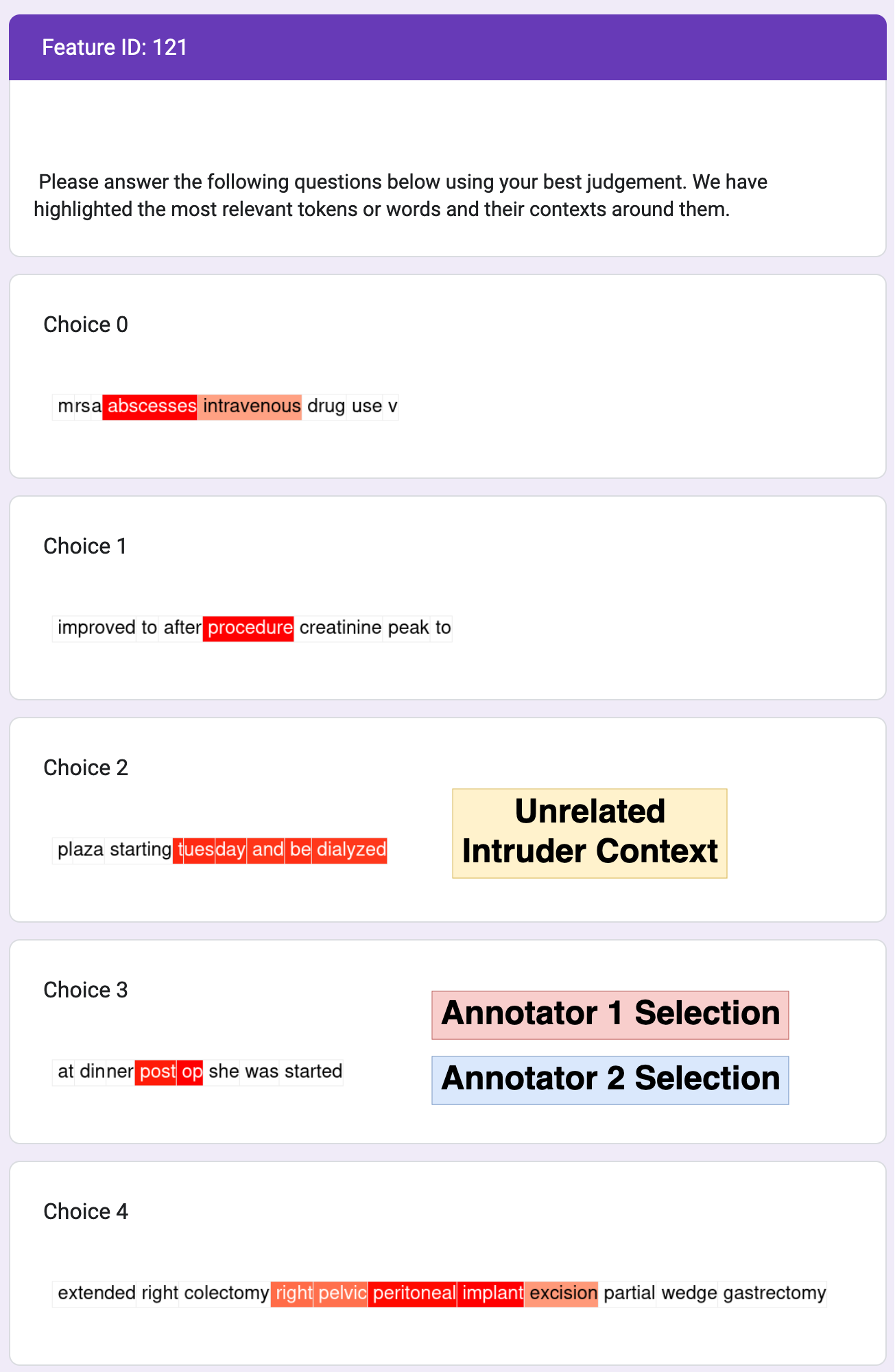}
\caption{\textbf{Uninterpretable $L_1$ feature neither annotator could differentiate:} Both annotators agreed that this dictionary feature lacked a distinctive theme.}
\label{fig:hard_l1_dict_feat2}
\end{figure}

\vspace{-0.2cm} 
We perform the word intrusion experiment from \citet{subramanian2017spine} for the sparse autoencoder features and other baselines. Figures \ref{fig:ez_l1_dict_feat}, \ref{fig:med_l1_dict_feat}, \ref{fig:med_l1_dict_feat2}, and \ref{fig:hard_l1_dict_feat2} showcase the expected unrelated "random context" in gold and the expert annotators' choices in red or blue for $L_1$ features. Qualitative examinations reveal varying levels of distinctiveness. Highly coherent features made it easy to distinguish the outlier token(s), but more complicated cases involving the level of abstraction of specific token contexts (Figure \ref{fig:med_l1_dict_feat}) were harder to discern, especially when the resolution of the token mattered. Other features were shown to be highly uninterpretable (Figures \ref{fig:med_l1_dict_feat2}, \ref{fig:hard_l1_dict_feat2}).


\begin{figure}[H]
\centering
\includegraphics[width=0.45\textwidth]{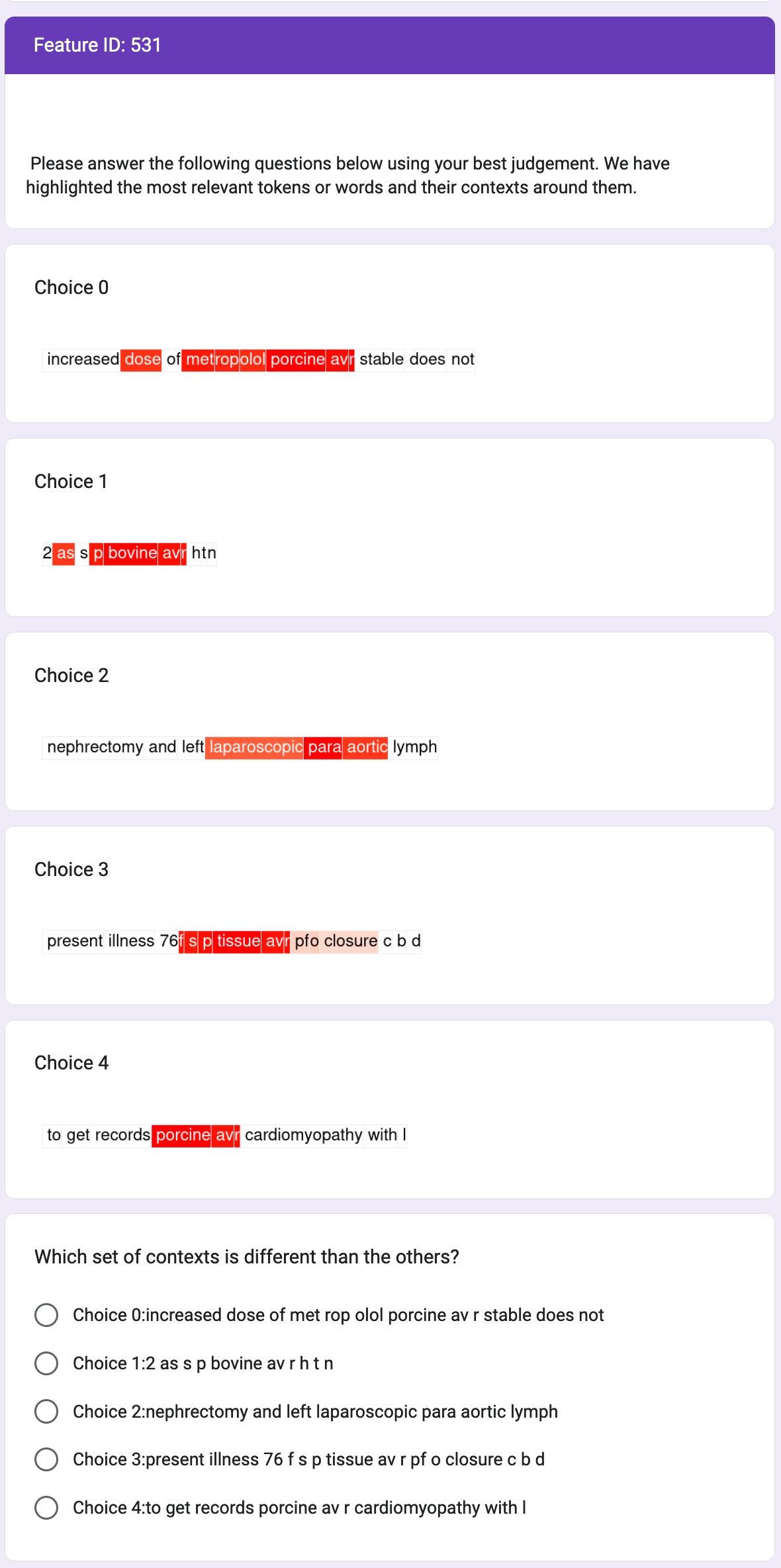}
\caption{Example of human evaluation form interface.}
\label{fig:exHumanEvalForm}
\end{figure}

\begin{figure}[H]
\centering
\includegraphics[width=0.45\textwidth]{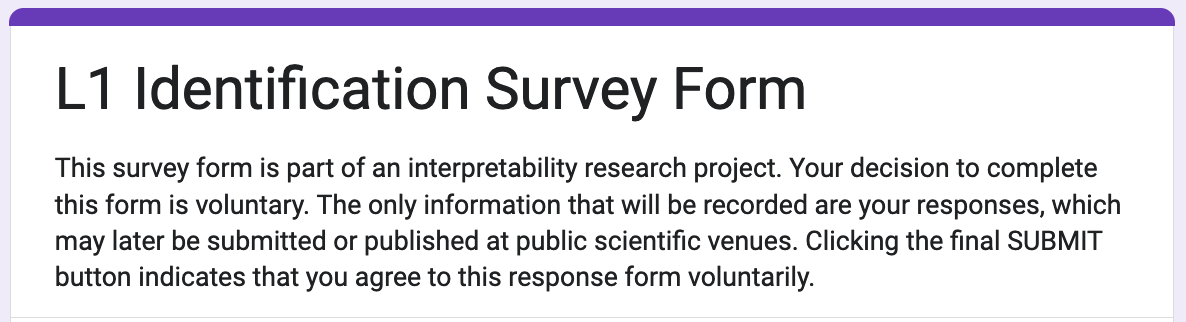}
\caption{Example of human evaluation form instructions.}
\label{fig:exHumanEvalFormInstructions}
\end{figure}
\newpage
\subsection{Initial Sparse Autoencoder Experiments}
We trained new sparse autoencoders for this revision, as we felt the previous ones were ill-trained due to not filtering out pads during the training process, leading to many irrelevant medical features. Surprisingly, despite finding that only approximately 500 features accounted for many of the ablation downstream results, their feature ablations still highly affected downstream performance, as shown in Table \ref{tab:ablation_old_gt}. Note that these experiments measured the top ground truth medical code rather than the top predicted medical code, hence the discrepancy in the probability drops. However, the order of these results remains the same. We also report the previous results, such as coherence and similarity, in Section \ref{appendix:prev_human_understand}. We were surprised to find the relative order of baselines in terms of explainability performance remained the same in this revision.
\begin{table*}[h!]
\centering
\resizebox{1.0\textwidth}{!}{%
\begin{tabular}{c|c c |c c c c c | c c |c c c c}
\toprule
\multicolumn{13}{c}{\textbf{Ablating Dictionary Features of Highlighted Tokens}} \\
\midrule
\multicolumn{1}{c|}{\textbf{Experiment}} & \multicolumn{2}{c|}{\textbf{ AutoCodeDL }} & \multicolumn{5}{c|}{\textbf{LAAT + Baselines }} & \multicolumn{2}{c}{\textbf{ DL }}  & \multicolumn{4}{|c}{\textbf{ Baselines }} \\
\midrule
 & \textbf{L1} & \textbf{SPINE} & \textbf{ICA} &  \textbf{PCA} & \textbf{Identity} & \textbf{Random}  & \textbf{Token} & \textbf{L1} & \textbf{SPINE} & \textbf{ICA} &  \textbf{PCA} & \textbf{Identity} & \textbf{Random} \\
\midrule
\textbf{Top}    & 0.685 & 0.686 & 6.169e-5  & 0.678 & 0.465 & 0.709 & 0.678 & 0.906  &0.929 & 0.001 & 0.884 & 0.744 & 0.939  \\
\textbf{NGT}  & 4.847 & 4.666 & 5.624 & 4.926 & 5.195 & 618.140 & 4.925   & 51.647 & 39.096 & 5.614 & 36.403& 493.625 & 370.035  \\
\textbf{Ratio} & \textbf{0.141} & \textbf{0.147} & 1.100e-5 & 0.138 & 0.090 & 0.001 & 0.138 & 0.0176 & 0.0238 & 1.344e-4 & 0.024 & 0.002 & 0.003   \\
\bottomrule
\end{tabular}%
}
\caption{Softmax probability changes in downstream ICD predictions resulting from ablation experiments. 'Top' represents the magnitude of softmax drops for the most probable ground truth ICD code, while 'NGT' signifies the sum of absolute softmax probability changes of non-ground truth ICD codes for each clinical note. The 'Ratio' indicates the ratio between these two measures. We bold and distinguish the results obtained using our combined LAAT and dictionary learning framework, and observe that our method has the most precise effect on downstream ICD predictions, suggesting improved explanatory power.}
\label{tab:ablation_old_gt}
\end{table*}

\begin{table}[h]
\centering
\resizebox{0.35\textwidth}{!}{%
\begin{tabular}{@{}lc|cccc@{}} 
\toprule
\multicolumn{6}{c}{\textbf{Hidden Medical Meaning Identification Accuracy}} \\ 
\midrule
\multicolumn{2}{c|}{\textbf{ AutoCodeDL }} & \multicolumn{4}{c}{\textbf{ Baselines}} \\
\midrule
\textbf{L1} & \textbf{SPINE} & \textbf{ICA} & \textbf{PCA} & \textbf{Identity} & \textbf{Random} \\
\midrule
0.794 & \textbf{0.864} & 0.351 & 0.388 & 0.327 & 0.297 \\
\bottomrule
\end{tabular}
}
\caption{Proportion of stop word embedding labels correctly identified by our AutoCodeDL framework using previous sparse autoencoders with using 99th percentile activated dictionary features, alongside the baseline methods. Such results showcase that DL is capable of effectively identifying hidden meanings embedded within superposition of stop words.
}
\label{table:StopWordExperimentsOld}
\end{table}

\subsubsection{Other Initial Ablation Experiments} \label{appendix:more ablation studies}
\begin{figure}[ht]
\centering
\includegraphics[width=0.45\textwidth]{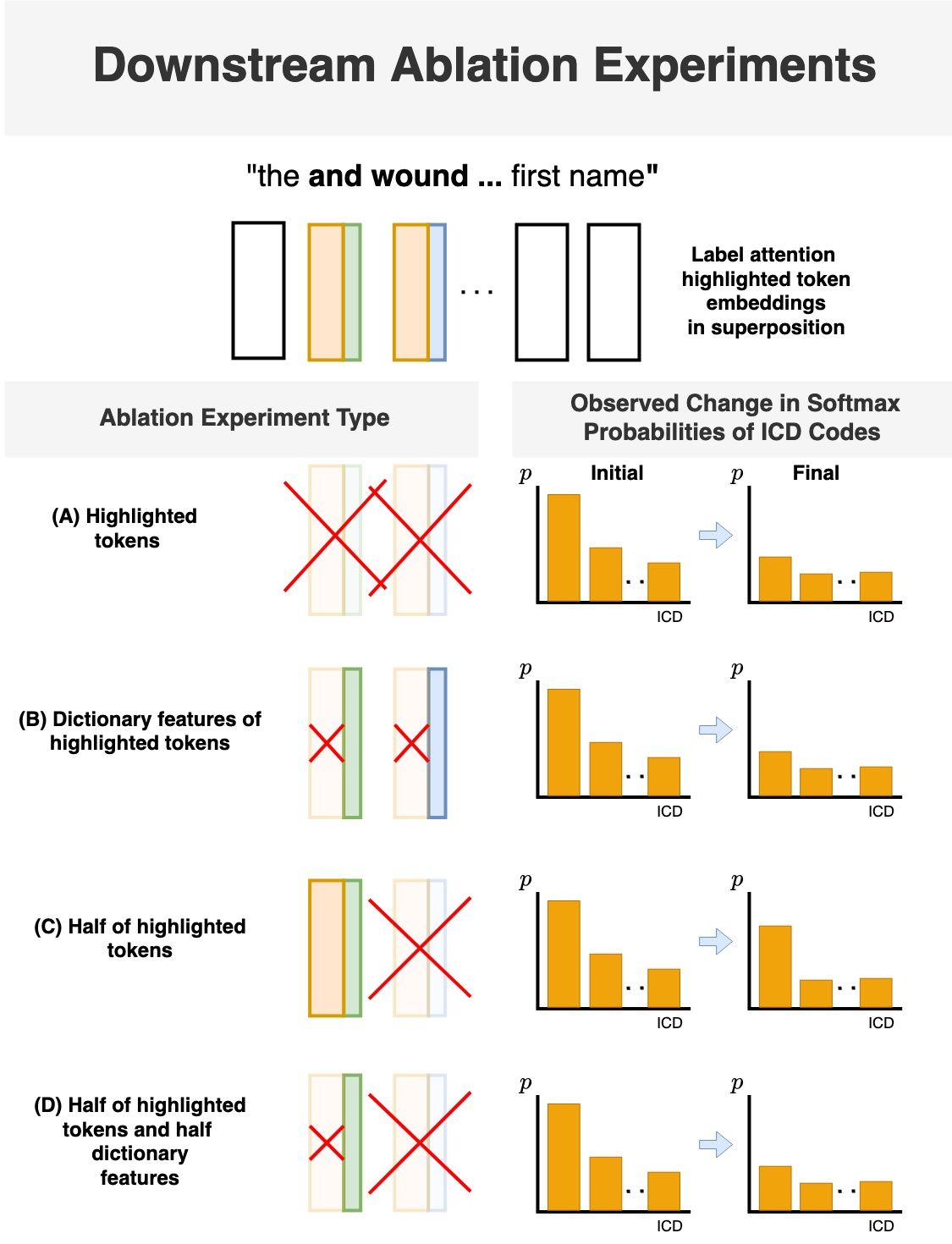}
\caption{Ablation experiments on clinical notes using a label attention (LAAT) mechanism to highlight relevant tokens for ICD coding: (A) Complete ablation of all highlighted tokens, or ablation of only the most relevant dictionary features of each token embedding (B), and (C) random ablation of half the highlighted tokens compared to (D) random ablation of half the highlighted tokens and the most relevant dictionary features of other half of token embeddings. }
\label{fig:ablation}
\end{figure}

We perform additional validation experiments to showcase that sparse autoencoder (dictionary learning) do outperform their none-sparse baselines in terms of model explainability. For reference, we have investigated the downstream effects of dictionary learning through a total of four ablation experiments (Figure \ref{fig:ablation}). Our analysis centered on the ICD code most likely to be predicted for each note, determined by softmax probabilities, and employed two token ablation benchmarks: (A) complete ablation of all highlighted tokens and (C) random ablation of half the top highlighted tokens. In parallel, dictionary features underwent ablation in two forms: (B) solely ablation of the paramount dictionary feature for all highlighted tokens or (D) ablating half of the tokens and ablating the most significant dictionary feature of the remaining highlighted tokens. For reference, tokens surpassing the 95th percentile in attention scores for their respective ICD code in the label attention matrix were considered "highlighted". We note that we have only shown experimental results for ablation types (A) and (B) in section \ref{sec:ExpResults}.

 To investigate if such ablations are perfectly additive, we perform further experiments (C) and (D) where we only ablate half of the relevant tokens and observe the overall changes from ablating the dictionary features of the unablated other half of tokens in Table \ref{tab:ablation_additive}.
\begin{table}[h!]
\centering
\resizebox{0.5\textwidth}{!}{%
\begin{tabular}{c|c c |c  c c c c}
\toprule
\multicolumn{1}{c|}{\textbf{Experiment}} & \multicolumn{2}{c|}{\textbf{(C) DL}} & \multicolumn{5}{c}{\textbf{(D) Baselines}} \\
\cmidrule{1-8}
\textbf{Measure} & \textbf{L1} & \textbf{SPINE} & \textbf{ICA} &  \textbf{PCA} & \textbf{Identity} & \textbf{Random} & \textbf{Tokens} \\
 \cmidrule{1-8}
\textbf{Top}   & 0.506 & 0.506 & 0.0519 & 0.205 & 0.068 & 0.743 & 0.0519 \\
\textbf{NGT}   & 5.146 & 5.112 & 5.294 & 5.164 & 5.281 & 11.876 & 5.296 \\
\textbf{Ratio}   & \textbf{0.098} & \textbf{0.099} & 0.0098 & 0.040 & 0.013 & 0.063 & 0.001 \\
\bottomrule
\end{tabular}%
}
\caption{Softmax probability changes in downstream ICD predictions resulting from ablation experiments illustrated in Figure \ref{fig:ablation}. 'Top' represents the magnitude of softmax drops for the most probable ground truth ICD code, while 'NGT' signifies the sum of absolute softmax probability changes of non-ground truth ICD codes for each clinical note. The 'Ratio' indicates the ratio between these two measures. We highlight and distinguish the results obtained using sparse autoencoders.}
\label{tab:ablation_additive}
\end{table}

We observe that while ablating such dictionary features do indeed drop the softmax probabilites of the most likely ICD code, they do not sum to the original softmax drop observed in the ablation experiment in section \ref{sec:ExpResults}, and thus there is some missing information that is not entirely encompassed by the dictionary feature ablations. That being said, DL still provides the best model explanation for downstream ICD predictions in this scenario.

\subsubsection{Initial Human Understandability Evaluations} \label{appendix:prev_human_understand}
\textbf{Coherence of top 500 activating dictionary features.} Utilizing Siamese encoder embeddings, we calculate the average cosine similarity among the top $k$ tokens of the top 500 dictionary feature to gauge their conceptual relatedness. For more methodological specifics of our Siamese BERT experiment, please refer to Appendix~\ref{appendix:siamese_cosine}. While not using the same sparse autoencoders, these results are still useful from at least a reproducibility standpoint. 

\textbf{Results.} Overall, the coherence of the unsupervised dictionary learning methods, represented by the DL columns in the table, decreased as the top k tokens considered increased. Contrary to expectations outlined by \cite{subramanian2017spine}, sparse autoencoders trained through $L_1$ minimization exhibited the highest coherence among all methods, as indicated by the highest average cosine similarity values across different values of $k$. This suggests that the highest activating dictionary features learned through $L_1$ minimization are more semantically consistent and conceptually coherent compared to other unsupervised methods.

Surprisingly, the coherence of dictionary features learned from the top independent components in ICA surpassed that of SPINE, despite ICA's lack of impact on downstream ICD coding performance. This unexpected finding highlights the complex interplay between feature extraction methods and semantic coherence.

As expected, the supervised LAAT method achieved the highest coherence, reflecting its use of labeled data to guide the learning process that results in more tokens closely aligned with the semantics of ICD codes.

\begin{table}[h!]
\centering
\resizebox{0.5\textwidth}{!}{%
\begin{tabular}{@{}lcc|cccc|cc@{}}
\toprule
\multicolumn{8}{c}{\textbf{Coherence of Top 500 Activated Dictionary Features}} \\ 
\midrule
\multicolumn{1}{c}{}  & \multicolumn{2}{c|}{\textbf{DL}} & \multicolumn{4}{c|}{\textbf{Baselines}} & \multicolumn{1}{c}{\textbf{Supervised}} \\
\midrule
$k$ &\textbf{L1} & \textbf{SPINE} & \textbf{ICA} &\textbf{PCA} & \textbf{Identity} &\textbf{Random} & \textbf{LAAT} \\ 
\midrule
\textbf{2} & \textbf{0.344} & 0.283 & 0.312 & 0.244 & 0.269 &0.271 & 0.692\\
\textbf{4} & 0.331 & 0.278 & 0.292  &  0.234 & 0.242 &0.251 & 0.678  \\
\textbf{10} & 0.303 & 0.260 & 0.280  & 0.228 & 0.234 & 0.238 & 0.637 \\
\bottomrule
\end{tabular}
}
\caption{Average cosine similarity between the top k tokens extracted from each of the top 500 dictionary feature or ICD code, measured from Siamese encoder embeddings. Higher values indicate a stronger thematic connection within the feature or code. The "DL" columns represents our dictionaries constructed, while the remaining columns are baselines.}
\label{table:tokens_compared_old}
\end{table}
\textbf{Dictionary ICD Overlap.} The descriptions of a dictionary feature's most pertinent ICD codes should logically coincide with its most highly activating tokens. Each ICD code comes with a description enriched with medically relevant information, suggesting that a dictionary feature that accurately encapsulates a medical concept will exhibit an overlap between the descriptions of its relevant ICD codes and its activating tokens. 

\textbf{Setup.} To assess this, we extract descriptions for each dictionary feature's relevant ICD codes, removing stop words, to compile a list of medically significant tokens. We calculate the overlap between these tokens and each medically relevant feature's top activating tokens, using the proportion of overlapping tokens as an additional measure of a dictionary feature's understandability. We define a feature to be medically relevant when its ablation results in at least a 10\% softmax probability drop of any ICD code. 

\textbf{Results.} Comparing our dictionary features' top tokens to those identified by specialized "label attention" (trained to map tokens to ICD codes), we find significant overlap for L1 and SPINE (Table \ref{table:OverlapBetweenICD9AndTopRelevantWords}). This overlap surpasses baselines, suggesting our unsupervised methods effectively learn medically relevant concepts. Notably, ICA features show minimal overlap. Overall, our dictionaries exhibit strong agreement with medically significant concepts, boosting our model's interpretability.

\begin{table}[h!] 
\centering
\resizebox{0.5\textwidth}{!}{%
\begin{tabular}{lcc|cccc|c} 
\toprule
\multicolumn{8}{c}{\textbf{Percentage of Dictionary Features GPT3.5 Differentiated }} \\ 
\midrule
\multicolumn{1}{c}{}  & \multicolumn{2}{c|}{\textbf{DL}} & \multicolumn{4}{c|}{\textbf{Baselines}} & \multicolumn{1}{c}{\textbf{Supervised}} \\
\midrule
 & \textbf{L1} & \textbf{SPINE} & \textbf{ICA} & \textbf{PCA} & \textbf{Identity} & \textbf{Random} & \textbf{LAAT} \\
\midrule
& 0.352 & \textbf{0.420} & 0.416 & 0.320 & 0.356 & 0.328 & 0.540  \\
\bottomrule
\end{tabular}
}
\caption{Percentage of the 500 randomly sampled dictionary features successfully distinguished by GPT3.5 Turbo, determined by selecting the unrelated token from a set of four tokens.}
\label{table:GPT3.5exp_old}
\end{table}

\begin{table}[h!]
\centering
\resizebox{0.5\textwidth}{!}{%
\begin{tabular}{@{}lcc|cccc|c@{}}
\toprule
\multicolumn{8}{c}{\textbf{Dictionary Overlap with ICD Descriptions}} \\ 
\midrule
\multicolumn{1}{c}{}  & \multicolumn{2}{c|}{\textbf{DL}} & \multicolumn{4}{c|}{\textbf{Baselines}} & \multicolumn{1}{c}{\textbf{Supervised}} \\
\midrule
 &\textbf{L1} & \textbf{SPINE} & \textbf{ICA*} & \textbf{PCA} & \textbf{Identity} & \textbf{Random} & \textbf{LAAT}\\
\midrule
& \textbf{0.117} & 0.086 & 0.000 & 0.010 & 0.024 & 0.003 & 0.198  \\
\bottomrule
\end{tabular}
}
\caption{Proportion of highly activating tokens in dictionary features that overlap with ICD9 descriptions. LAAT represents an upper bound in overlap where we treat each ICD row in the label attention matrix as its own dictionary feature. *In general, the ablation of the independent components in ICA have little effect on downstream ICD predictions, hence its features have no overlap with ICD descriptions in this experiment.}
\label{table:OverlapBetweenICD9AndTopRelevantWords}
\end{table}

\subsection{Initial Examples of Dictionary Features} \label{appendix:ex_dictionary_features}
\begin{figure*}[h!]
\centering
\includegraphics[width=0.7\textwidth]{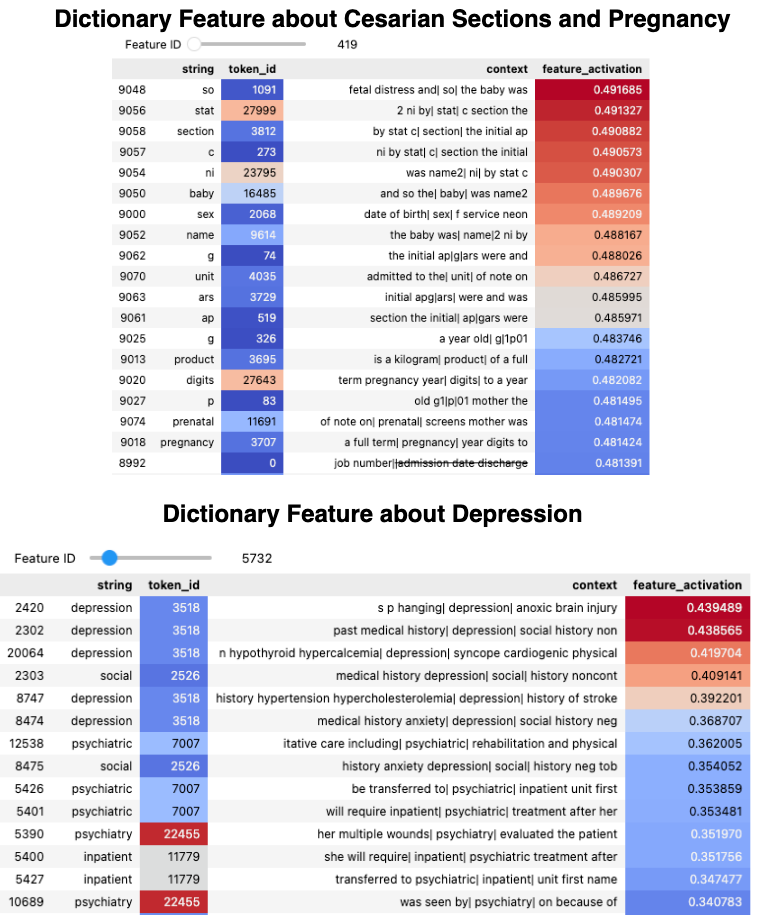}
\caption{Examples of dictionary features. We showcase the most relevant tokens for dictionary feature about cesarian sections and depression as well as their clinical note contexts. We note that these sorted pandas dataframes are from an earlier sparse autoencoder in the previous revision of the paper. However, we felt that they were still worth showcasing. }
\label{fig:DictExFeatures}
\end{figure*}

\begin{figure*}[h!]
\centering
\includegraphics[width=0.7\textwidth]{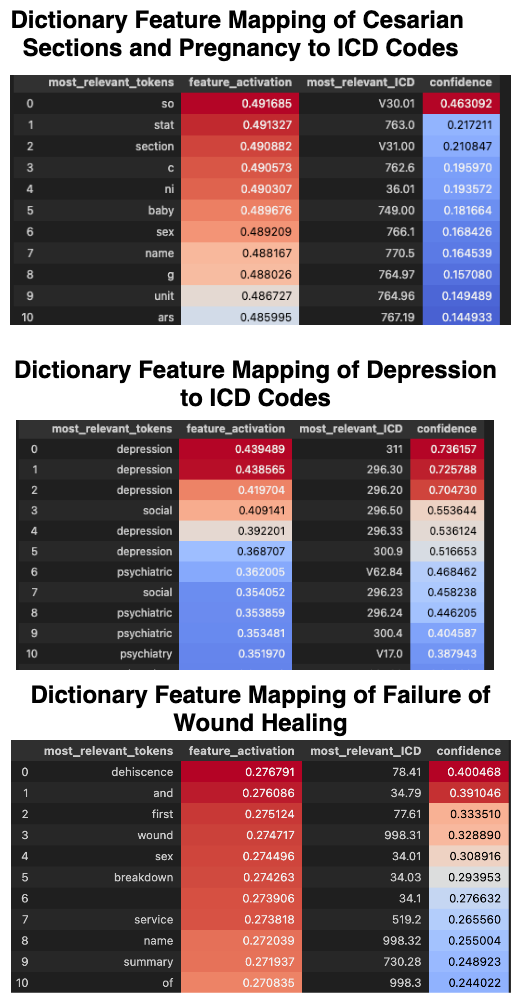}
\caption{Examples of dictionary features. We showcase the most relevant tokens and ICD codes for each respective interpretable dictionary feature. We note that these sorted pandas dataframes are from an earlier sparse autoencoder trained in the previous revision of the paper. However, we felt that they were still worth showcasing to prove a point. They contain various medical codes that are directly related to each token. }
\label{fig:DictExFeatures2}
\end{figure*}

We manually inspect several highly interpretable dictionary features as a showcase of the potential of these interpretable representations. We show some examples in Figure \ref{fig:DictExFeatures} and \ref{fig:DictExFeatures2}, relating to depression, cesarian sections, and failure of wound healing.

